\documentclass[pdftex,twocolumn,epjc3]{svjour3}          

\RequirePackage[T1]{fontenc}

\smartqed  

\RequirePackage{graphicx}
\RequirePackage{mathptmx}      
\RequirePackage{flushend}
\RequirePackage[numbers,sort&compress]{natbib}
\RequirePackage[colorlinks,citecolor=blue,urlcolor=blue,linkcolor=blue]{hyperref}

\usepackage{bm}

\usepackage{placeins}

\usepackage[utf8]{inputenc} 
\usepackage{url}            
\usepackage{booktabs}       
\usepackage{nicefrac}       
\usepackage{microtype}      
\usepackage{array}
\usepackage[caption=false]{subfig}
\usepackage[dvipsnames]{xcolor}

\expandafter\let\csname equation*\endcsname\relax

\expandafter\let\csname endequation*\endcsname\relax

\usepackage{amsmath}

\usepackage{floatrow}
\usepackage{bbm}

\usepackage{algorithm}
\usepackage[noend]{algpseudocode}
\usepackage{xparse}
\makeatletter
\NewDocumentCommand{\LeftComment}{s m}{%
  \Statex \IfBooleanF{#1}{\hspace*{\ALG@thistlm}}\(\triangleright\) #2}
\makeatother

\usepackage{caption}
\DeclareCaptionFormat{algor}{%
  \hrulefill\par\offinterlineskip\vskip1pt%
    \textbf{#1#2}#3\offinterlineskip\hrulefill}
\DeclareCaptionStyle{algori}{singlelinecheck=off,format=algor,labelsep=space}
\journalname{Eur. Phys. J. B}

\begin{document}

\title{Belief propagation for supply networks: Efficient clustering of their factor graphs}

\author{Tim Ritmeester\thanksref{e1,addr1}
        \and
        Hildegard Meyer-Ortmanns\thanksref{e2,addr1} 
}

\thankstext{e1}{e-mail: t.ritmeester@jacobs-university.de}
\thankstext{e2}{e-mail: h.ortmanns@jacobs-university.de}

\institute{Physics and Earth Sciences,
  Jacobs University Bremen,  P.O. Box 750561, 28725 Bremen, Germany \label{addr1}
}

\date{}

\abstractdc{We consider belief propagation (BP) as an efficient and scalable tool for state estimation and optimization problems in supply networks such as power grids. BP algorithms make use of factor graph representations, whose assignment to the problem of interest is not unique. It depends on the state variables and their mutual interdependencies. Many short loops in factor graphs may impede the accuracy of BP. We propose a systematic way to cluster loops of naively assigned factor graphs such that the resulting transformed factor graphs have no additional loops as compared to the original network. They guarantee an accurate performance of BP with only slightly increased computational effort, as we demonstrate by a concrete and realistic implementation for power grids. The method outperforms existing alternatives to handle the loops. We point to other applications to supply networks such as gas-pipeline or other flow networks that share the structure of constraints in the form of analogues to Kirchhoff's laws. Whenever small and abundant loops in factor graphs are systematically generated by constraints between variables in the original network, our factor-graph assignment in BP complements other approaches. It  provides a fast and reliable algorithm to perform marginalization in tasks like state determination, estimation, or optimization issues in supply networks.}

\maketitle

\section{Introduction} \label{sec: introduction}
Belief Propagation (BP) is an algorithm that is known from statistical physics \cite{del_ferraro_cavity_2014,yedidia_understanding_2003}, computer science, artificial intelligence and information science (for a review see, for example, \cite{yedidia_message-passing_2011}).
It runs also under the name of `message passing'. 
In some cases BP provides an exact rearrangement of the original calculational objective, while in general it calculates a mean-field approximation \cite{yedidia_understanding_2003, del_ferraro_cavity_2014, yedidia_message-passing_2011,yedidia_constructing_2005} to that.
The BP method has two established main advantages over traditional algorithms (such as least-squares and quasi-Newton methods \cite{huang_state_2012, abeysekera_steady_2016}). It is fast even for large networks, as the computation time scales linearly in the system size, and it is robust against large differences in the input parameters.
Its robustness avoids convergence issues associated with the traditional approach (\cite{meliopoulos_distributed_2008}). These properties make it uniquely suitable for dealing with large datasets and for frequent and large-scale network analyses (if possible online), which are increasingly required for supply networks.

In electric power grids, BP is in principle applicable to optimization problems and state estimation (in Section~\ref{sec:3} we give a concrete implementation for state estimation). Optimization problems refer to cost-efficient and low-risk performance under diverse sources of uncertainties. State estimation is the procedure of using measurement data to infer an estimate of state variables such as power flows and phase angles as accurately as possible. State estimation is important to convert system measurements into reliable information on the true network state and to ensure the stability of operation. In \cite{cosovic_state_2016, cosovic_observability_2020, ritmeester_state_2020} BP based algorithms were found to outperform traditional (least-squares) approaches to state estimation. In particular, the speed and robustness of the BP-based algorithms enable state estimation in real-time \cite{cosovic_state_2016,hu_belief_2011,weng_graphical_2013} and statistical analyses of large networks \cite{cosovic_observability_2020, ritmeester_state_2020}, even if data is partially missing \cite{ritmeester_state_2020}. Thus state estimation combined with BP amounts to an important step in view of improving supervisory control and planning decisions in running power grids if a fast online estimation of the true state of the grid is required.


A second range of applications of BP are gas networks that we consider in some more detail in the appendix.
Natural gas is still one of the important energy resources worldwide. A reliable and efficient operation of gas-pipeline networks becomes increasingly important due to the liberalization of the European gas market.
In \cite{abeysekera_steady_2016} the impact of injecting alternative gas supplies at different locations is studied to facilitate decisions on the allowable amount and composition also of alternative gas sources such as hydrogen and biogas.

BP has been applied to  further supply networks \cite{han_enabling_2018, ortega_contamination_2020, yeung_physics_2013, wong_inference_2007}, other than power or gas networks.
The authors of \cite{han_enabling_2018} and \cite{ortega_contamination_2020} used BP to identify faults and contamination sources in water networks.
In \cite{yeung_physics_2013} a BP algorithm was shown to effectively optimize public transport for urban planning and telecommunication networks, 
while \cite{wong_inference_2007} applied BP to a generic nonlinear resource allocation problem. \\
{}\\
For a given network with an associated graph, the BP algorithm makes use of an additional auxiliary graph called the factor graph. The novel contribution of this paper is the assignment of an appropriate factor graph to supply networks. If the very factor graph has a tree structure, BP is known to be exact \cite{tanner_recursive_1981}. In general, BP implements an approximation corresponding to Bethe mean-field theory
\cite{yedidia_understanding_2003, yedidia_message-passing_2011,yedidia_constructing_2005}. Practically, BP is  accurate if there are only a few short loops in the factor graph.\\
 A naively assigned factor graph directly reflects the basic variables on the original network and their mutual dependencies. In the supply networks that we consider in this paper these interdependencies result from Kirchhoff's laws, corresponding to the conservation of flow at the vertices of the network (first law) and the constraint that the drop in variables like the voltage or pressure sums up to zero around elementary loops (second law). In general, these laws  impose nonlinear constraints on the flows, and these constraints are responsible for additional loops in a naively assigned factor graph, even if the original network is a tree. It is these loops that we want to avoid by a suitable assignment of factor graphs. The loops are numerous as the constraints between the variables are omnipresent in the network. The mathematical structure of these constraints is found in many supply networks, so that our proposed new algorithm applies to all of them.
%
Although our method succeeds in avoiding the many additional small loops in the factor graphs which result from constraints as mentioned before, it does not address the possible challenges which result from loops in the original supply networks. In the concrete test cases that we consider, the resulting algorithm still shows an excellent performance without addressing these loops. In cases where these loops do impede the performance of BP, our method should be combined with additional approaches. Several such improvements have been proposed and implemented, see, for example \cite{cosovic_state_2016,HeskesConvergent,pretti_message-passing_2005,wemmenhove_loop_2007, Rizzo}.
Such a combination is possible because our method, which relies on a type of clustering (to be defined below), changes only the factor graph rather than the BP algorithm itself.
In contrast to other clustering methods, we propose a systematic way of clustering factor graphs  in terms of clusters which by construction depend only on a few variables. This way the price to pay for clustering remains moderate. The resulting method
then improves the speed and convergence of the BP algorithm.

The paper is organized as follows.
In Section~\ref{sec:2} we describe the clustering procedure, assigning a factor graph which differs from a straightforward assignment, but prevents by construction additional loops in the factor graph assignment.
In Section~\ref{sec:3} we illustrate the application of the procedure with state estimation for the artificial IEEE-300 electrical grid. We discuss the accuracy and performance of the algorithm.
In Section~\ref{sec:5} we give an outlook to other applications of our algorithm to power grids. The conclusions are summarized in Section~\ref{sec:6}. In \ref{sec:4} we point to further possible supply networks which  share the essential structure of the equations; in particular we work out the case of natural gas-pipeline networks with an example from the steady state analysis of two realistic GasLib benchmark networks. Here our method  enables the very applicability of BP, as with our method,
 BP converges exceedingly fast while BP with a naive factor graph assignment does not converge at all.

\section{Assigning factor graphs to supply networks}\label{sec:2}

We consider supply networks consisting of vertices and links and assume the following generic description:
\begin{itemize}
    \item To each link in the supply network,  a flow of some quantity is assigned which traverses the link; flows from vertex $i$ to vertex $j$ and vice-versa are denoted as  $f_{ij}$ and $f_{ji}$, respectively.
    \item Associated with each vertex $i$ is a  variable $v_i$; examples are the voltage in electric circuits, or the pressure in fluid networks.
    \item The flow $f_{ij}$ through a link $(ij)$ is determined by the variables $v_i$ and $v_j$ at the vertices at either end of the link (e.g., Ohms law in electric circuits). The flow is thus a function $f_{ij}(v_i,v_j)$ of the vertex variables $v_i$ and $v_j$.
    \item Flows are conserved at each vertex: The sums of in and outgoing flows are equal. Denoting external injections into a vertex $i$ as $g_i$, we thus have $g_i = \sum_{j \in N(i)} f_{ij}(v_i, v_j)$,
    where $N(i)$ is the set of vertices adjacent to vertex $i$. The injection can thus be written as a function $g_i(v_i, v_j: j \in N(i))$ of the vertex variables $\{v_i, v_j: j \in N(i)\}$.
\end{itemize}
The class of problems that we study on these supply networks is defined by the criterion that the state is determined in terms of a probability distribution $P(\bm{v})$, depending on the vertex variables $\bm{v}$ of the following form:
\begin{equation} \label{eq: basicP}
P(\bm{v})=\prod_i H_i(v_i, v_j:j\in N(i))\times\prod_{(ij)} H_{(ij)}(v_i,v_j),
\end{equation}
where the products run over all vertices $i$ and all links $(ij)$ of the network, respectively,
 and $N(i)$ is the set of vertices $j$ that are connected to $i$ via a link.
The function $H_i$ depends on the variables $\{v_i, v_j:j\in N(i)\}$, which can in particular incorporate a mutual dependence on $v_i$ and on the injection $g_i(v_i, v_j: j \in N(i))$.
The $H_{ij}$ are functions depending on pairs of variables $v_i,v_j$, which may incorporate a dependence on the flows $f_{ij}(v_i,v_j)$ and $f_{ji}(v_j,v_i)$.
In Section~\ref{sec:3} and~\ref{sec:4} we will study  the state estimation of power grids and a network analysis of gas pipelines,
where we will explicitly show that these problems reduce to the marginalization of a probability distribution of the form of Eq.~\ref{eq: basicP}.\\\\
{\bf Applications of Eq.~\ref{eq: basicP}.} Although Eq.~\ref{eq: basicP} may look rather specific, it is in fact a general formulation that comprises the following cases:
\begin{itemize}
    \item Uncertainty in the state variables: The probability distribution may express the uncertainty in the values of the vertex variables $v_i$, the flows $f_{ij}$ and/or the injections $g_i$.
          Such uncertainty may arise from fluctuating injections (for example, in renewable energy generation),
          or from uncertainties in the measurements of these quantities (as in the Bayesian state estimation problem studied in Section~\ref{sec:3}).
          (If the value of a quantity is certain, this can be incorporated as a delta function in the distribution. If nothing about a quantity is known, the associated factor is assigned  a very high variance.)
    \item Optimization problems: For a given cost function $C(\bm{v})$, one considers the distribution
        \begin{eqnarray} \label{eq: optimization_as_P}
        P_T(\bm{v})\propto \exp(-C(\bm{v})/T) \,.
        \end{eqnarray}
        In the limit $T \rightarrow 0$ the probability distribution peaks at the minimal costs $C(\bm{v})$.
        The exponential turns sums into products, such that the distribution of Eq.~\ref{eq: basicP}  incorporates a
        minimization of a sum of costs on the vertex variables, the flows and the injections.
        Constraints can be implemented by setting the cost function equal to infinity whenever the constraints are violated. It is possible to explicitly take this limit in the BP Eqs.~\ref{eq: BP1}-\ref{eq: BP3} below, thereby converting them to a form that is more convenient for optimization (called the min-sum algorithm).
        We refer to \cite{yedidia_message-passing_2011,dvijotham_graphical_2017} for details.
    \item Constraint satisfaction problems (i.e., finding configurations $\bm{v}$ that satisfy a number of constraints): These can be included by studying products of Dirac delta functions,
          where each delta function incorporates a constraint. Alternatively, a cost function is optimized that assigns a penalty to each violated constraint.
          We will use this option to analyze the steady state of gas-pipeline networks in \ref{sec:4}.
    \item Optimization under uncertainty: If costs need to be minimized in an inherently fluctuating environment  \cite{altarelli_stochastic_2011},
            decisions on the production, for example, may lead to stochastic rather than deterministic costs.
            For power grids such a situation has been investigated in \cite{summers_stochastic_2014, bienstock_chance-constrained_2014}, where fluctuations are due to uncertain power injections by renewable resources.
            \end{itemize}

If the distribution of Eq.~\ref{eq: basicP} should be evaluated to gain insight into the probability of individual variables, it amounts to a marginalization of this joint probability distribution by summing or integrating over a subset of variables. This is the place where BP enters in the sense that the sums or integrals are performed in a very efficient way.

\subsection{The choice of factor graphs} \label{sec: BP_basics}
BP can be used to efficiently calculate marginals of probability distributions such as those of Eq.~\ref{eq: basicP}.
It is convenient if BP makes  use of a graphical representation of the probability distribution in terms of a factor graph.
The factor graph is a bipartite graph, made of two types of nodes, variable nodes, represented by circles, and factor nodes, represented by squares.
The assignment of variables and factors is not unique and a matter of convenience.
The procedure of assigning a factor graph to any probability distribution $P(\bm{v})$ proceeds in the following steps:
\begin{enumerate}
    \item Partition the vector $\bm{v}$ into new 'variables' $\{x_{I}\}$, where each $x_I$ is a disjoint subset of $\bm{v}$ (i.e., possibly containing multiple $v_i$).
    \item For each $x_I$, draw a circle. This defines a \textit{variable node} of the factor graph. As a special case, the correspondence between variables on the original grid and the variable nodes on factor graphs may be one to one in a straightforward assignment (which we refer to as the naive assignment).
    \item Define factors $W_a(\bm{x}_a)$ such that $P(\bm{v}) = \prod_a W_a(\bm{x}_a)$, where $\bm{x}_a$ are (in general overlapping) sets of some of the new variables $\{x_I\}$.
    \item For each factor $W_a$, draw a square. This is a \textit{factor node} of a factor graph.
    \item Each factor $W_a(\bm{x}_a)$ depends on $\bm{x}_a$, which contains multiple variable nodes $x_I$.
          Draw an edge between the factor node $W_a(\bm{x}_a)$ and each variable node $x_I \in \bm{x}_a$ on which it depends.
\end{enumerate}
 Ambiguities in the assignment of the factor graph result from step (1) and (3).\\

\begin{figure}
    \centering
    \subfloat[]{\includegraphics[trim = 170 500 40 170, clip, width = 0.9 \textwidth]{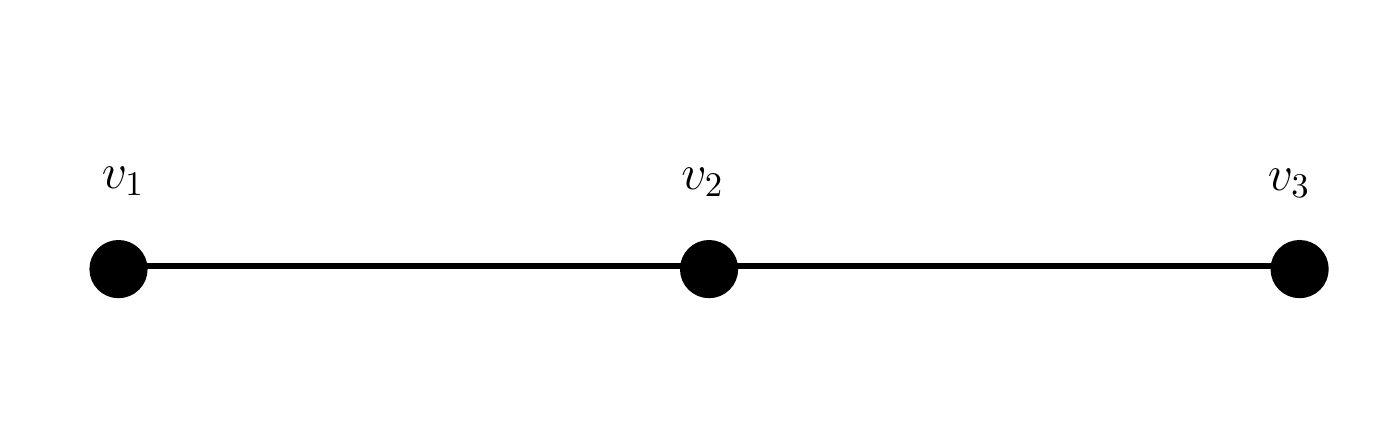}} \hspace{8cm}\\
    \subfloat[]{\includegraphics[trim = 0 310 0 0, clip, width = 0.9 \textwidth]{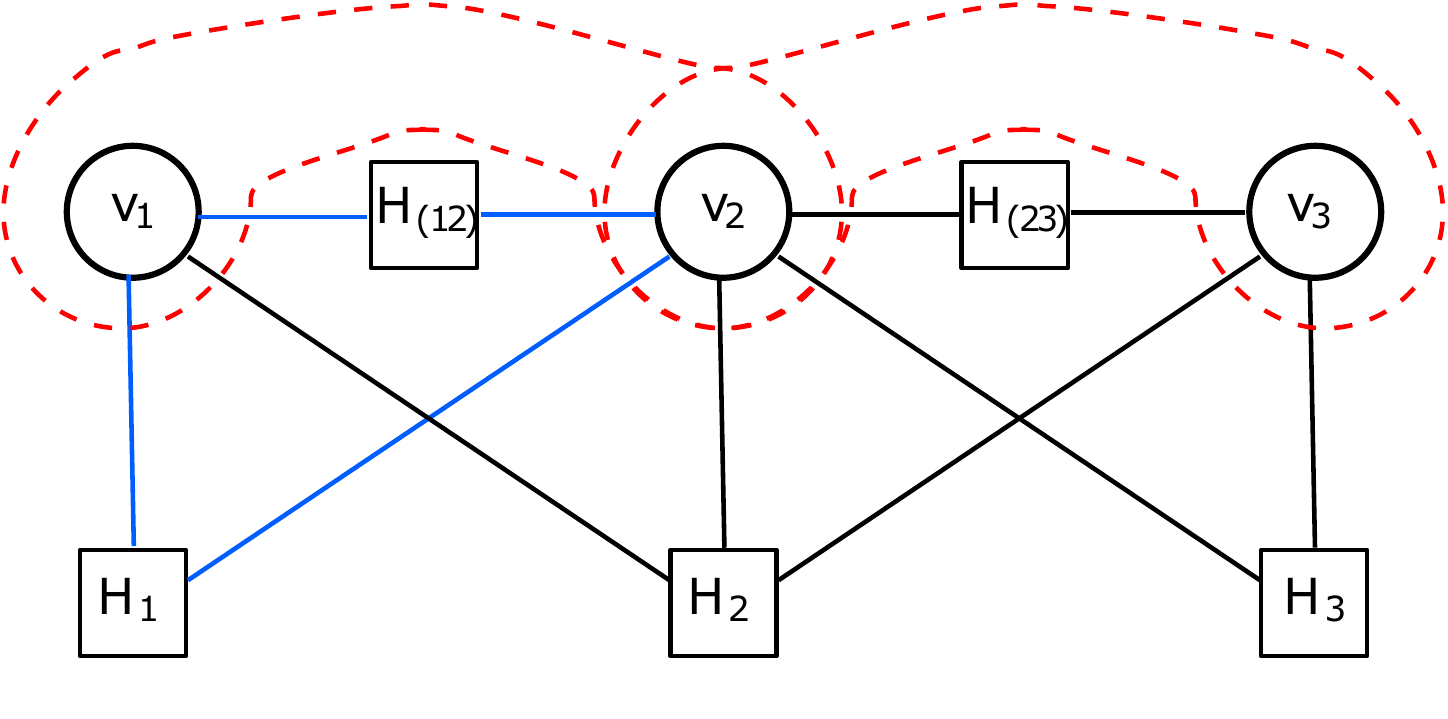}}
    \caption{(a) Simple building block of a supply network, which gives the probability distribution of Eq.~\ref{eq: example_P}. (b)~The factor graph assigned in a straightforward way (Section 2.1) to Eq.~\ref{eq: example_P}, representing the simple network shown in (a). Note that the four elementary loops (one of them indicated in blue) would be absent if each $H_i$ would depend only on $v_i$ rather than on all $\{v_i, v_j: j \in N(i)\}$. The dotted lines indicate two clusters of vertices, which the approach described in Section~\ref{sec: clustering} will use to eliminate the loops (after dealing with the overlap at vertex $v_2$).}
    \label{fig: simple_network}
\end{figure}

{\bf Example of a straightforward choice of a factor graph.} 
Let us give a concrete example, considering a simple building block  of three vertices $\{1,2,3\}$ of a larger network, connected by links $(1,2)$ and $(2,3)$ (shown in Fig.~\ref{fig: simple_network}(a)).
The distribution (Eq.~\ref{eq: basicP}) we are interested in is then given by:
\begin{eqnarray} \label{eq: example_P}
    P(v_1,v_2,v_3) &= &\big[H_1(v_1,v_2) \cdot H_2(v_1,v_2,v_3) \cdot H_3(v_2,v_3) \big] \nonumber \\
                    	        &	\times &\big[H_{(1,2)}(v_1,v_2) \cdot H_{(2,3)}(v_2,v_3) \big] \,.
\end{eqnarray}
Following the steps (1)-(4), the most straightforward way of assigning a factor graph to this distribution
is to assign a variable node to each $v_i$, and a factor node to each $H_i$ and each $H_{(ij)}$.
The factor graph corresponding to the distribution (\ref{eq: example_P})
is shown in Fig.~\ref{fig: simple_network}(b).

In particular, it is important to note that this straightforward assignment  has a number of loops (four in this case, one of them indicated in blue) though the original network (Fig.~\ref{fig: simple_network}(a)) has none.
Responsible for the loops are the factors $H_i$: Without these factors, the topology of the factor graph would directly reflect the topology of the original supply network (for each vertex it contains a variable node $v_i$, and for each link it contains a factor node $H_{ij}$ connecting variables nodes $v_i$ and $v_j$). The clustering applied to this factor graph (to be discussed later) refers to the variable nodes, in Fig.~\ref{fig: simple_network}(b) we indicate two clusters with dotted lines, overlapping in variable node $v_2$. In Fig.~\ref{fig: clusters_tree} below we show how to deal with the overlaps.
Moreover, if $H_i$ would depend only on $v_i$ (rather than  on all  $\{v_i, v_j: j \in N(i)\}$), the factor nodes corresponding to $H_i$ would be leaf nodes and  create no loops.\\
{}\\
{\bf Number of extra loops in a naive assignment.} The total number of extra loops can be determined as follows.
In general, the total number of loops in a graph, here the factor graph, is given by ($ \sharp\;\text{connected components} + \sharp\;\text{edges} \\ - \sharp\;\text{vertices} $) \cite{harary_graph_2001}.
The naively assigned factor graph contains factor nodes $H_i$ and $H_{ij}$ and variable nodes $v_i$ as well as edges connecting these nodes.
As mentioned before, if we consider only the variable nodes, the factor nodes $H_{ij}$ and the edges connecting them,
the resulting structure directly reflects the topology of the original supply network,  in particular it has the same amount of loops. The amount of extra loops in the factor graph (as compared to the supply network) can thus be calculated as
($\sharp\;\text{additional edges due to factor node}\;\{H_i\} - \sharp\;\text{additional nodes which are of type}\;H_i$).
On the factor graph, each $H_i$ must be connected by an edge to all variable nodes in $\{v_i, v_j: j \in N(i) \}$:
The amount of extra edges is thus given by $\sum_i |v_i, v_j: j \in N(i)| = \sum_i(1 + |N(i)|)$. The number of extra nodes is simply $\sum_i \,1$ (one factor node for each $H_i$).
In total the number of extra loops in the factor graph is thus given by $\sum_i (1 + |N(i)|) - \sum_i 1 = \sum_i |N(i)| = 2\cdot\sharp \; \mbox{links}$.
This means in the example of Fig.~\ref{fig: simple_network}(b) that there are $2 \cdot 2 =4$ extra loops (where the number of links can be found from Fig.~\ref{fig: simple_network}(a)).
The dependence of $H_i$ on further variables from $N(i)$ due to ubiquitous constraints on all of the network variables thus leads to a proliferation of loops in the factor graph.
\subsection{Sketch of the BP algorithm} \label{sec: BP_algorithm}
To fix the notation, we summarize the basic steps of the BP algorithm. For a given factor graph with variable nodes $\{x_I\}$ and factor nodes $W_a(\bm{x}_a)$ (such that $P(\bm{x}) = \prod_a W_a(\bm{x}_a) $), BP can be used to calculate marginals $P_I(x_I) \equiv \int \prod_{J \neq I}\mathrm{d} x_J P(\bm{x})$ and $P_a(\bm{x}_a) \equiv \int \prod_{J: x_J \notin \bm{x}_a}\mathrm{d} x_J P(\bm{x})$. Here we will give the basic BP-algorithm, for which several extensions exist  \cite{cosovic_state_2016,HeskesConvergent,pretti_message-passing_2005,wemmenhove_loop_2007, Rizzo}, as mentioned in the introduction.
The steps of the BP-algorithm for the marginalization are the following:\\
{\it Initialization.}
For each factor-variable pair $(a,I)$  that is connected on the factor graph (that is, for which $x_I \in \bm{x}_a$), messages $\{m_{I \rightarrow a}(x_I), m_{a \rightarrow I}(x_I)\}$ are initialized uniformly:
At time $t=0$, the messages are set to $m_{I \rightarrow a}^{t = 0}(x_I) \propto 1$ and $m_{a \rightarrow I}^{t = 0}(x_I) $ \\$ \propto 1$. The messages are functions of the variables, If the variables are discrete or if the factors are Gaussian (implying also Gaussian messages) the messages can be parameterized by a few real numbers. Otherwise one needs to find an approximation, such as a discretization of the messages \cite{dvijotham_graphical_2017,NIPS2008_b73ce398} or the basis function expansions considered in \cite{noorshams_low-complexity_nodate}. In case of Gaussian distributions we initialize the messages with zero mean and large variance to generate a uniform distribution. \\
{\it Updates.}
At each step $t$, we keep track of approximations $\{b_I^t(x_I)\}$ and $\{b^t_a(\bm{x}_a)\}$ (with $I$ and $a$ running over all variable and factor nodes respectively), which for large $t$ are supposed to converge to the marginals of $P(\bm{x})$ according to $b_I(x_I)\rightarrow P_I(x_I)$ and $b_a(\bm{x}_a)\rightarrow P_a(\bm{x}_a)$.
The approximations are given in terms of the messages $\{m_{I \rightarrow a}^t(x_I), m_{a \rightarrow I}^t(x_I) \}$, which are updated according to:
\begin{eqnarray}
    m_{J \rightarrow a}^{t+1}(x_J) &=& \prod_{b \neq a; x_J \in \bm{x}_b} m_{b \rightarrow J}^t(x_J) \label{eq: BP1} \\
    m_{a \rightarrow I}^{t+1}(x_I) &=& \int \big[ \hspace{-0.3cm} \prod_{J \neq I; x_J \in \bm{x}_a}  \hspace{-0.3cm} m_{J \rightarrow a}^{t+1}(x_J) \big] \times  \big[W_a(\bm{x}_a) \big] \times  \hspace{-0.3cm}  \prod_{J \neq I; x_J \in \bm{x}_a}   \hspace{-0.3cm} \mathrm{d}\, x_J \,, \label{eq: BP2}\\
    b_I^{t+1}(x_I) &\propto& \prod_{a: x_I \in \bm{x}_a} m_{a \rightarrow I}^{t+1}(x_I) \,. \label{eq: BP3}\\
    b_a^{t+1}(\bm{x}_a) &\propto& \prod_{a: x_I \in \bm{x}_a} m^{t+1}_{I \rightarrow a}(x_I) \,.\label{eq: BP4}
\end{eqnarray}
Note that the notation $x_I \in \bm{x}_a$ means that variable node $I$ and factor node $a$ are connected on the factor graph. The updates are repeated until a reasonable stopping criterion is reached, for example, when $b^{t+1}_I(x_I) - b^{t}_I(x_I)$ or $b^{t+1}_a(\bm{x}_a) - b^{t}_a(\bm{x}_a)$ reach some desired tolerance.
\\
{\it Output.}
The sets $\{b_I(x_I)\}$ and $\{b_a(\bm{x}_a)\}$ then give the approximations of the marginals $\{P_I(x_I)\}$ and $\{P_a(\bm{x}_a)\}$, respectively, that we want to determine.

\subsection{Systematic clustering of short loops in factor graphs} \label{sec: clustering}
Importantly, in relation to our application to supply networks, we should distinguish between the topology of the original network such as the power grid and the topology of the associated factor graph.
As we have seen in Section 2.1,
even if the  graphical representation of the original network is a tree, a naively assigned factor graph may unavoidably contain loops. Our statements below refer always to the topology of the factor graph.

Clustering is a method to find factor graph representations with a reduced number of loops by aggregating multiple variable nodes and/or factor nodes into a reduced set of variable or factor nodes.
For variable nodes this corresponds to treating a subset of the variables as a new single variable node,
while for factor nodes their clustering corresponds to multiplying factors together to obtain a new, aggregated factor.
The catch in the choice of factor graphs is that the larger the subsets $\bm{x}_a$,
the more difficult is the calculation of messages resulting from Eqs.~\ref{eq: BP1}-\ref{eq: BP2}.
In the extreme case, where the whole distribution is clustered into a single factor,
the algorithm simply returns the original marginalization problem $b_I(x_I) = \int P(\bm{x}) \prod_{J \neq I} \mathrm{d} x_J$, such that BP is exact but  of  no advantage anymore.
Thus it is important to find a clustering that 1. guarantees a high accuracy, and 2. avoids difficulties in computing the messages via  Eqs.~\ref{eq: BP1} - \ref{eq: BP2}.
\begin{figure*}
    \centering
    \hspace{-0.3cm}\subfloat[]{\includegraphics[trim = 200 300 100 150, clip, width = 0.272 \textwidth]{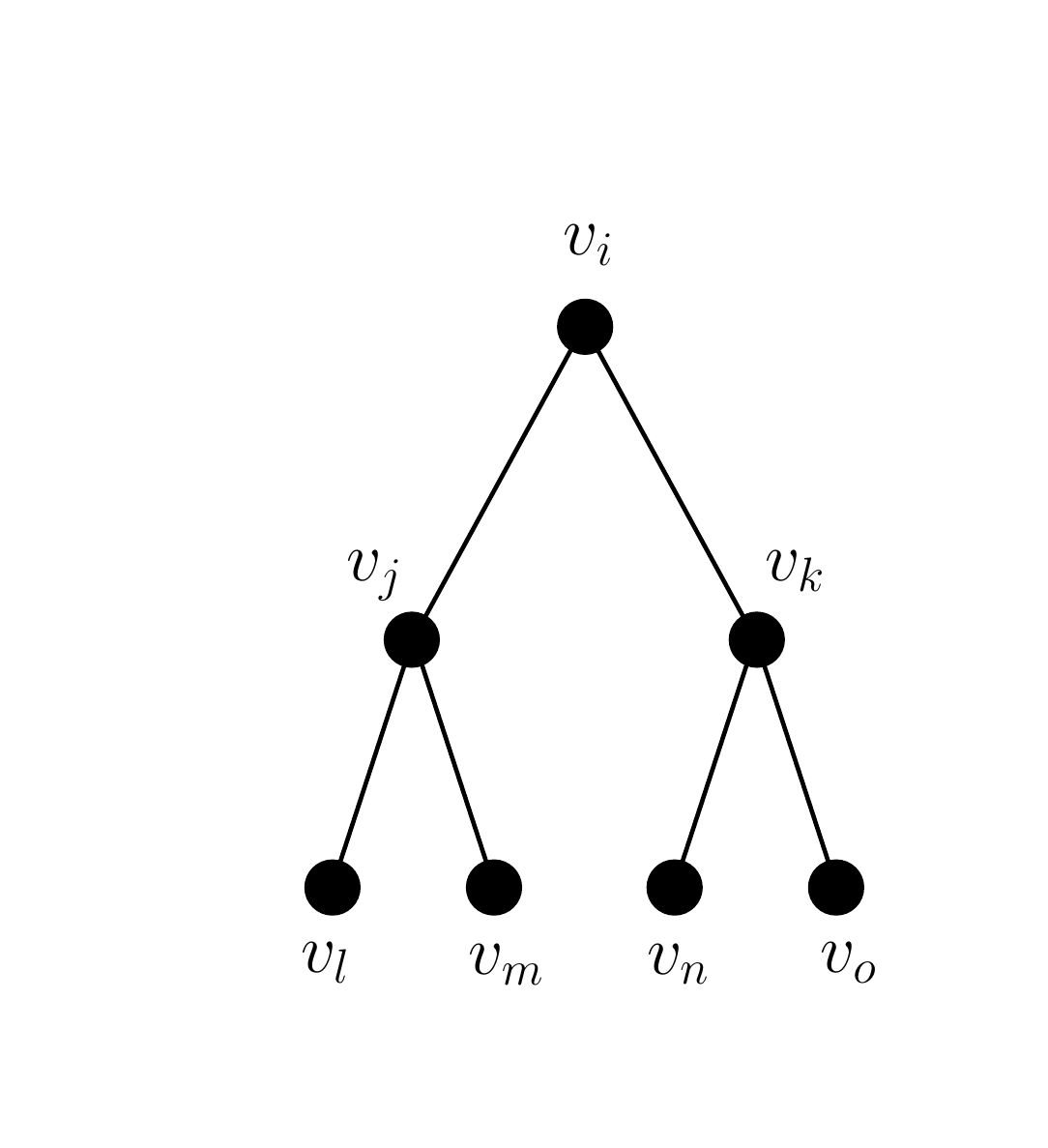}}
    \hspace{2.5cm}
    \subfloat[]{\includegraphics[trim = 200 300 100 150, clip, width = 0.272 \textwidth]{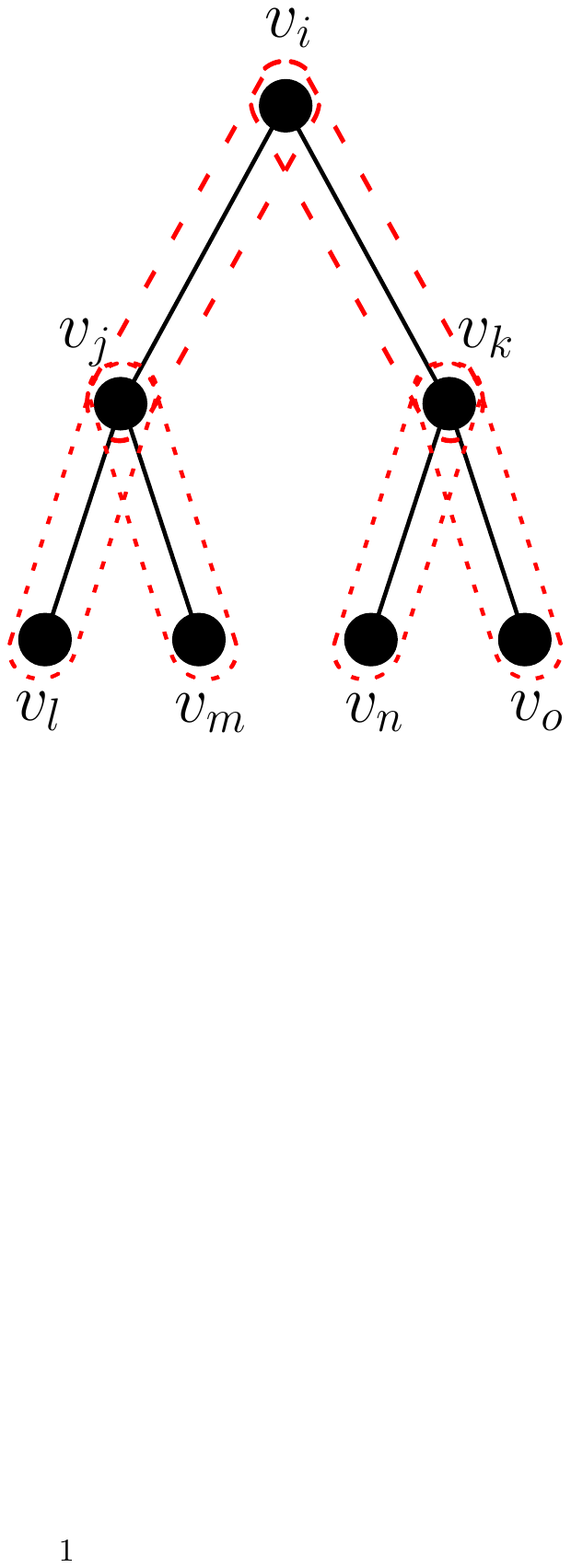}}\\
    \vspace{-0.3cm}
    \hspace{0.2cm}
    \subfloat[]{\includegraphics[trim = 240 300 105 200, clip, width = 0.32 \textwidth]{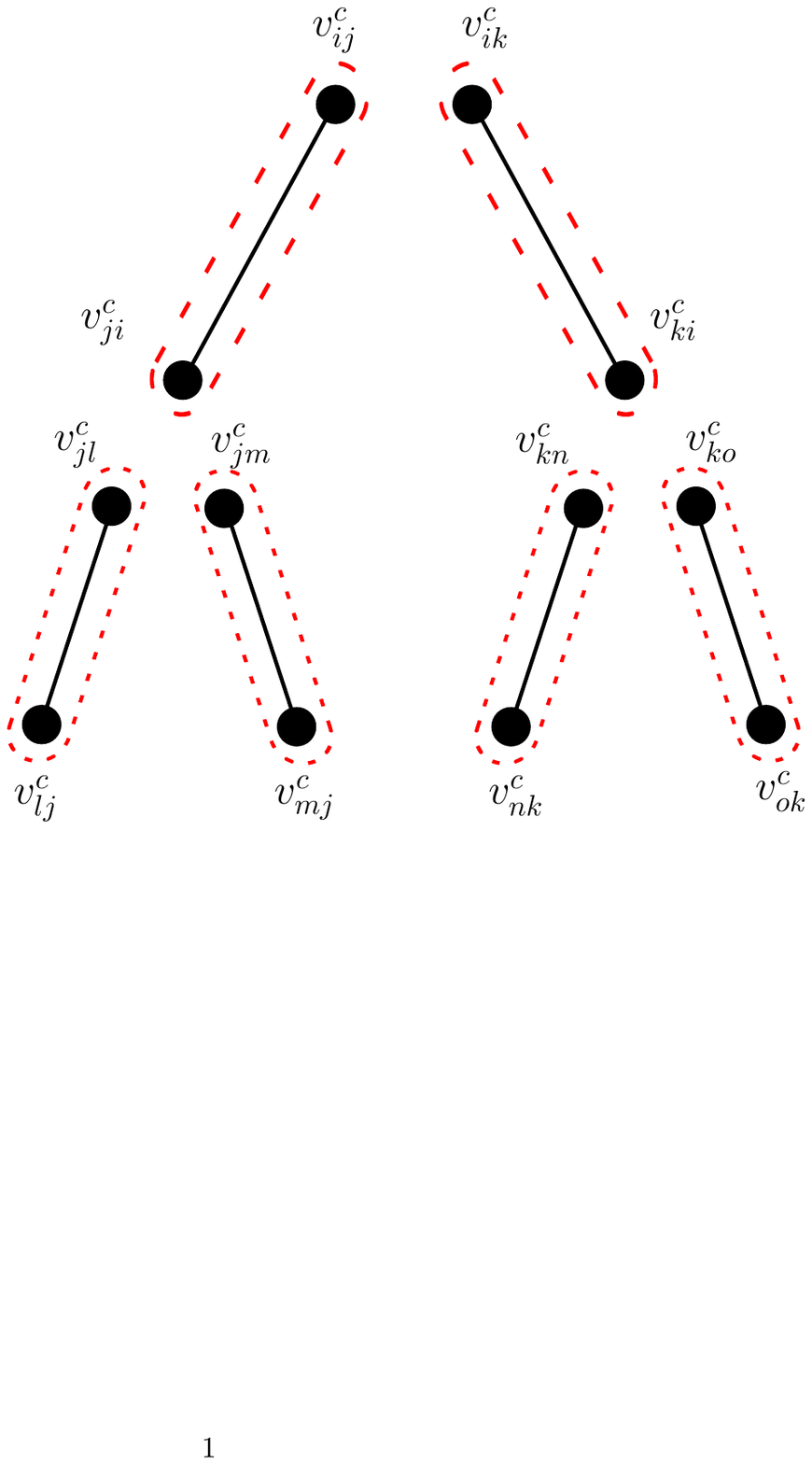}}
    \hspace{1.5cm}
    \subfloat[]{\includegraphics[trim = 240 300 105 200, clip, width = 0.32 \textwidth]{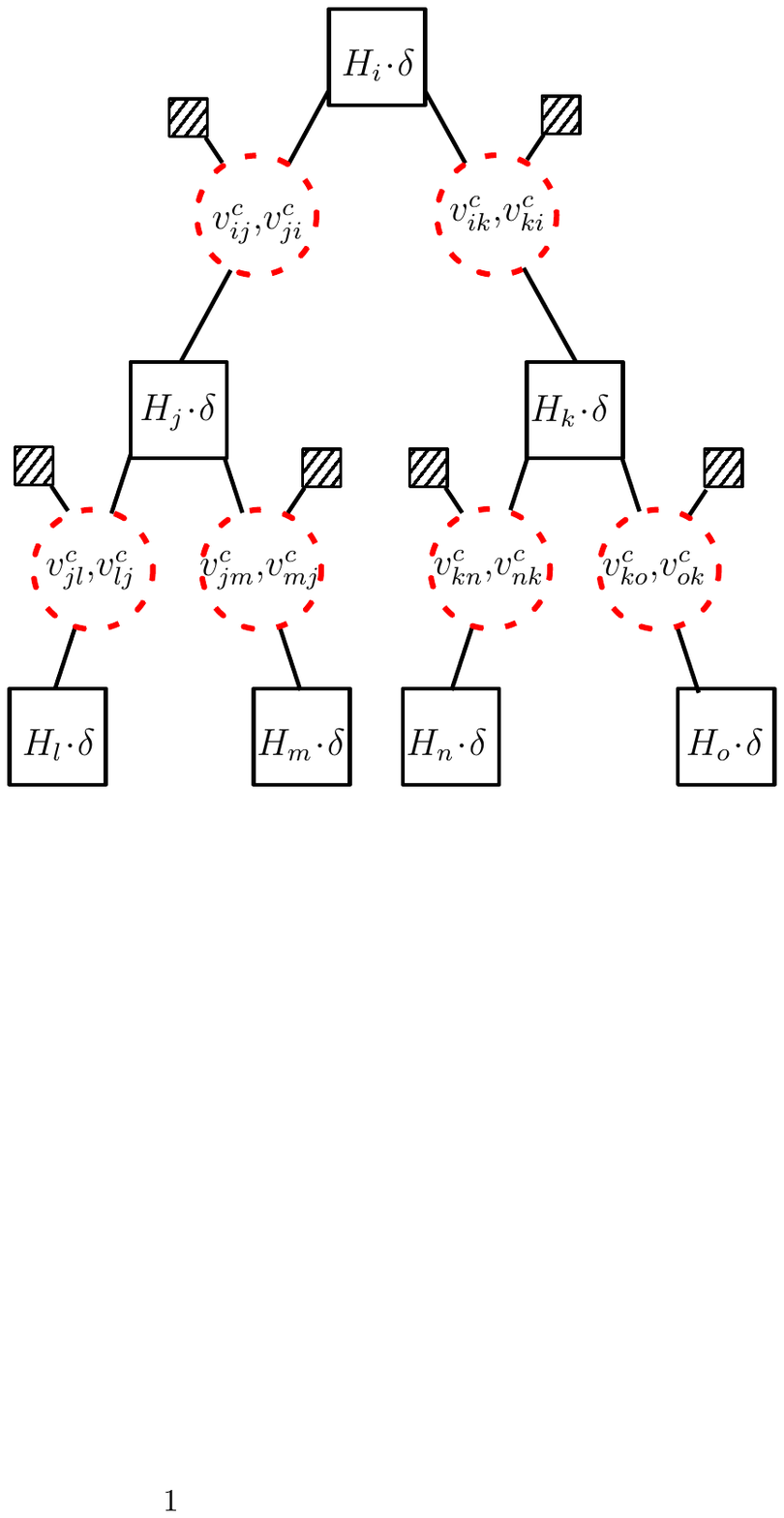}}
    \caption{(a) A simple tree network. (b) The striped ellipses indicate the clusters proposed as new variable nodes: Each cluster is assigned to a link and consists of the vertex variables at each end of that link. (c) Copying process of the vertex variable nodes on the factor graph to avoid overlapping. (d) Adding factor nodes $H_i$ between those variable nodes that enter $H_i$ of Eq.~\ref{eq: Pc} below, including the $\delta$-constraints, and attaching leaf nodes (dashed squares) (one for each variable node) that represent the $H_{ij}$-terms of Eq.~\ref{eq: Pc}. Note that (d) has the same tree structure as (a), where the links of (a) with attached vertices become variable nodes in (d) and the vertices of (a) have their counterpart in $H_i,\delta$-factor nodes with edges correspondingly attached.
    The links of the original network furthermore give an additional leaf node on the factor graph.}
    \label{fig: clusters_tree}
\end{figure*}
\subsubsection{Generating the clustered factor graph}\label{sec: clustered_graph}
In the clustered factor graph that we propose, we assign a variable node to each link of the original network.
Thus, each of these variable nodes is a tuple consisting of the vertex variables at each end of that link.
For a simple tree network, as in Fig.~\ref{fig: clusters_tree}(a),  this is indicated in Fig.~\ref{fig: clusters_tree}(b).
In Fig.~\ref{fig: clusters_tree}(b) the tuples of vertex variables are shown that are supposed to make up the variable nodes of the clustered factor graph.
For each link $(ij)$, the variable node at this link consists of the two vertex variables $(v_i,v_j)$ at each end of the link. However, the variable nodes are overlapping in the sense that each vertex variable $v_i$ is contained in multiple variable nodes: The variable node $(v_i, v_j)$ for each link $(ij): j \in N(i)$ connected to vertex $i$ contains the variable $v_i$. When copying the vertex variables to decouple the clusters (Fig.~\ref{fig: clusters_tree}(c)),
one has to compensate the copying by introducing $\delta$-constraints which enforce that all copies of a given vertex variable remain equal. This copying procedure is a generalization of the Shafer-Shenoy algorithm \cite{shenoy_axioms_1990} (in the sense that the Shafer-Shenoy algorithm requires the so-called 'running intersection property', while the copying procedure here does not). Mathematically this amounts to an identity operation, but it allows us to improve the performance of BP by eliminating loops from the factor graph.
Connecting the clustered variable nodes to the factor nodes then leads to a tree-like factor graph as in Fig.~\ref{fig: clusters_tree}(d).

\begin{figure}
   \centering
   \includegraphics[trim =145 300 70 360, clip, width = 1. \textwidth]{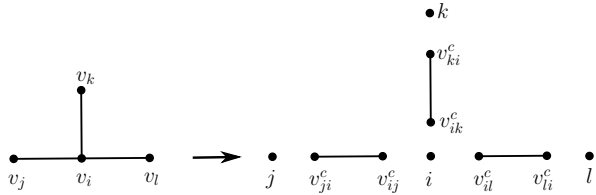}
   \caption{Copying procedure of the vertex variables as already used in Fig.~\ref{fig: clusters_tree}. The variable $v_i$ is copied to $v^c_{ij}, v^c_{ik}$ and $v^c_{il}$, one copy for each incident link.}
   \label{fig:copies}
\end{figure}

The number of clusters a given vertex variable belongs to, is equal to the number of links connected to the vertex. We thus have to make a copy for each of those links.
Thus we define a probability distribution which depends on all the copies of $v$-variables, while the original $v$-variables are integrated out by delta constraints:
\begin{eqnarray} \label{eq: shafer-shenoy_cluster}
    P_c(\bm{v}^c)  \equiv  \int P(\bm{v}) \prod_{i} \Big(  \Big[ && \prod_{j \in N(i)} \delta(v^c_{ij} - v_i) \Big] \mathrm{d} v_i   \Big) \,, \nonumber
\end{eqnarray}
where $\bm{v}^c \equiv \{v^c_{ij}: j \in N(i)\}$ is the set of all copies of vertex variables $v_i$ kept for those links $(ij)$ that are connected to vertex $i$ (thus, $v_{ij}^c$ is a copy of $v_i$).
Stated differently, each link connecting vertices $i$ and $j$ keeps a copy of $v_i$ and $v_j$ at its ends, and the delta function constrains the copies to remain equal to the original variables.
Fig.~\ref{fig:copies} shows the multiplication of vertex $i$ by three further copies, carried by the incident links toward vertices $j,l,k$.
From the definitions it is clear that calculating marginals in $P_c$ is equivalent to calculating marginals in the original distribution $P(\bm{v})$. The advantage is that $P_c$ contains no extra loops, it can be represented as a loop-free factor graph for which BP is exact if the supply network itself has no loops.
Writing out $P_c$ explicitly, we get:
\begin{eqnarray} \label{eq: Pc}
    P_c(\bm{v}^c) &=& \Big[ \prod_{i} \int \mathrm{d} v_i \, \Big\{H_i(v_i, v^c_{ji}: j \in N(i)) \prod_{j \in N(i)} \delta(v^{c}_{ij} - v_i) \Big\}\Big] \nonumber \\ &\times& \Big[ \prod_{(i,j)} H_{(ij)}\big(v^c_{ij}, v^{c}_{ji}) \big) \Big] \,.
\end{eqnarray}
In the first product we have explicitly kept the $v$-integration to include the $\delta-$constraints, while the $v$-integration has been carried out in the second product. Note that each $H_{ij}$ depends on a single variable on the new factor graph, while $H_i$ may depend on several variables on the factor graph.\\
Now we are ready to define the clustered factor graph.
As anticipated already in Fig.~\ref{fig: clusters_tree} (a)-(d), we are  able to assign a factor graph to $P_c$ which has the same amount of loops as the original network.\\
{}\\
\textbf{Variable nodes.} As variable nodes we use the tuples $(v^c_{ij}, v^c_{ji})$, one for each link $(ij)$ of the original network (Fig.~\ref{fig: clusters_tree}(a)).\\
\textbf{Factor nodes.}  Each vertex $i$ of the original network gives a factor node  $\int \mathrm{d} v_i \, \Big\{H_i(v_i, v^c_{ji}: j \in N(i)) \prod_{j \in N(i)} \delta(v^{c}_{ij} - v_i) \Big\}$, abbreviated as $H_i\cdot \delta$, see Fig.~\ref{fig: clusters_tree}(d).
Each link $(ij)$ of the original network gives a factor node $H_{(ij)}\big(v^c_{ij}, v^{c}_{ji})$. \\
{}\\
The new factor graph is illustrated in Fig.~\ref{fig: clusters_tree}(d).
The new variable nodes partition $\bm{v}^c$. Multiplying all the factors together gives the full distribution $P_c(\bm{v}_c)$ of Eq.~\ref{eq: shafer-shenoy_cluster}.
This is thus a valid factor graph. Note that here the new variable nodes $(v^c_{ij}, v^{c}_{ji})$ are  assigned to the links rather than to the vertices of the original network.
The reason is that the copied variables enter always in pairs, one vertex variable for each end of the link. The original variables can be retrieved from the corresponding copied variables at any of the links entering the vertex from the various directions.\\

\noindent \textit{Claim: The resulting factor graph has exactly as many loops as the original supply network.}
The factor at each vertex $i$ depends on $\{(v^c_{ij}, v^c_{ji}): j \in N(i) \}$; thus on the factor graph we connect the factor nodes $(H_i\cdot\delta)$ at each vertex $i$ to all variables $(v^c_{ij}, v^c_{ji})$ representing links incident to $i$.
Given only variables $(v^c_{ij}, v^c_{ji})$ and factors at the vertices of the supply network,
the factor graph thus has the same topology as the supply network,
with a variable for each link and a factor for each vertex, see Fig.~\ref{fig: clusters_tree}(d).
The other factors, associated with each link, depend on only one variable node each and are thus leaf nodes.
The topology of the factor graph thus equals the topology of the supply network with the addition of some leaf nodes.
In particular, it therefore has the same number of loops.
This is  graphically seen in Fig.~\ref{fig: clusters_tree}(d).

\subsubsection{The flow-only factor graph as a special case.} \label{sec: flow-only}
Before comparing the results of BP on the straightforward and the clustered factor graphs (Section~2.1 and Section~\ref{sec: clustered_graph}), let us first discuss a special case of our clustering.
The clustering simplifies if the whole distribution (Eq.~\ref{eq: basicP}) can be written in terms of flows $f_{ij}(v_i,v_j)$ rather than vertex variables:
\begin{equation} \label{eq: Pflow}
    P_f(\bm{f}) \equiv \Big[ \prod_{i} H_i(\sum_{j \in N(i)}f_{ij})\Big] \times \Big[ \prod_{(i,j)} H_{ij}(f_{ij}) \Big] \,,
\end{equation}
and additionally we assume that $f_{ij} = -f_{ji}$ to implement flow conservation. 
If one would explicitly keep the dependence on $\bm{v}$, writing the flows as $f_{ij}(v_i,v_j)$, one sees that the distribution still corresponds to a distribution of the form of Eq.~\ref{eq: basicP}.
If we assign a factor graph to this distribution $P_f$ in a straightforward way, using as variables $f_{ij}$ ($= - f_{ji}$) for each $(ij)$, and factors $H_i$ for each vertex and $H_{ij}$ for each link,
the factor graph has exactly the same topology as the clustered factor graph we proposed. The advantage is that the expressions are simpler due to the absent dependence on $v_i,v_j$.

Not all distributions of the form of Eq.~\ref{eq: basicP} can be written as in Eq.~\ref{eq: Pflow} in terms of flow only.
For example, the distribution of Eq.~\ref{eq: Pflow} cannot constrain the flows to Kirchoff's second law, so one requirement is that the original network is a tree. In this case the flows are exclusively determined by the conservation law $g_i = \sum_{j \in N(i)} f_{ij}$,
and  do not receive extra constraints from the vertex variables $\bm{v}$. Another requirement is that  the choice of the distribution of Eq.~\ref{eq: basicP} does not involve the vertex variables $\bm{v}$ directly,
but only indirectly through $\{g_i\}$ and $\{f_{ij}\}$.
If the network does not satisfy these requirements, the flow-only distribution (Eq.~\ref{eq: Pflow}) can still be used as an approximation,
where we ignore the constraints that the vertex variables $\bm{v}$ induce on the flows. In the case of an electric power grid,  this corresponds to ignoring Kirchoff's second law, taking only power flow conservation into account. We call this approximation the ``flow-only''-approximation, which we considered in \cite{ritmeester_state_2020}.
We will further compare this approximation to our clustering in Section~\ref{sec:3}.

\subsubsection{Comparison of factor graphs}
\begin{figure*}
    \centering
    \subfloat[]{\includegraphics[trim = 170 400 40 200, clip, width = 0.45 \textwidth]{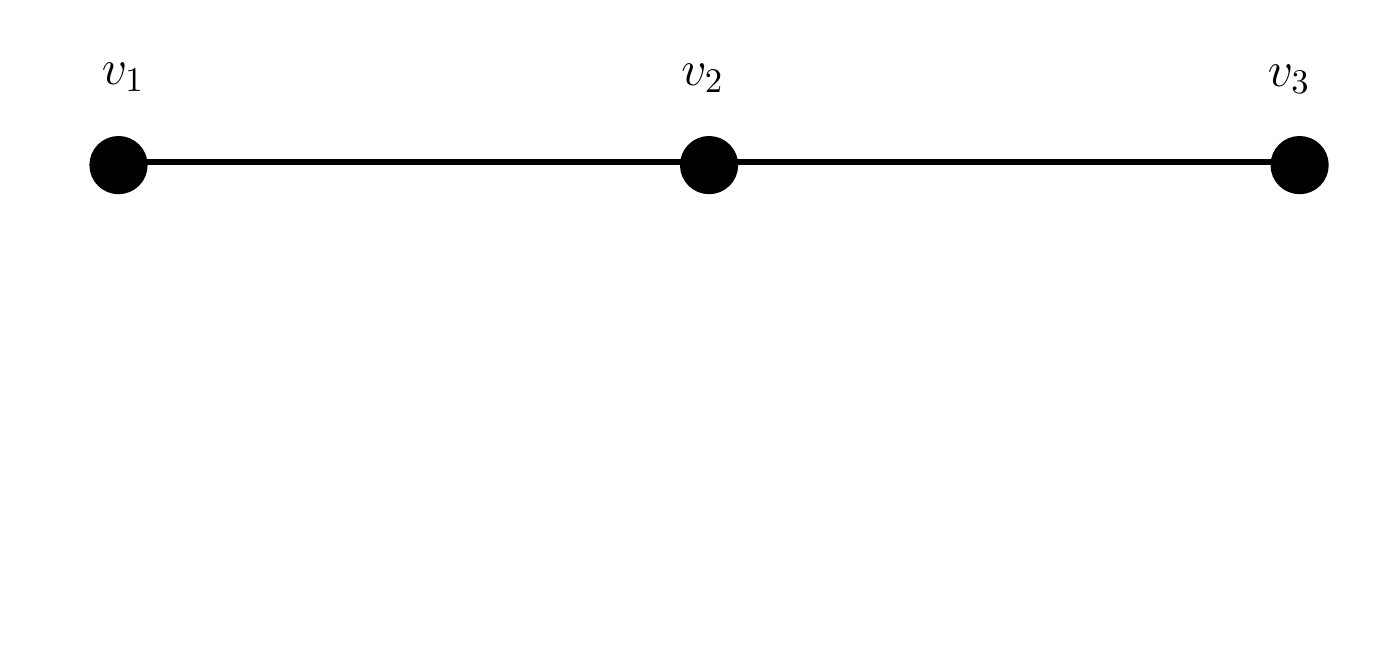}}
    \subfloat[]{\includegraphics[trim = 170 400 40 200, clip, width = 0.45 \textwidth]{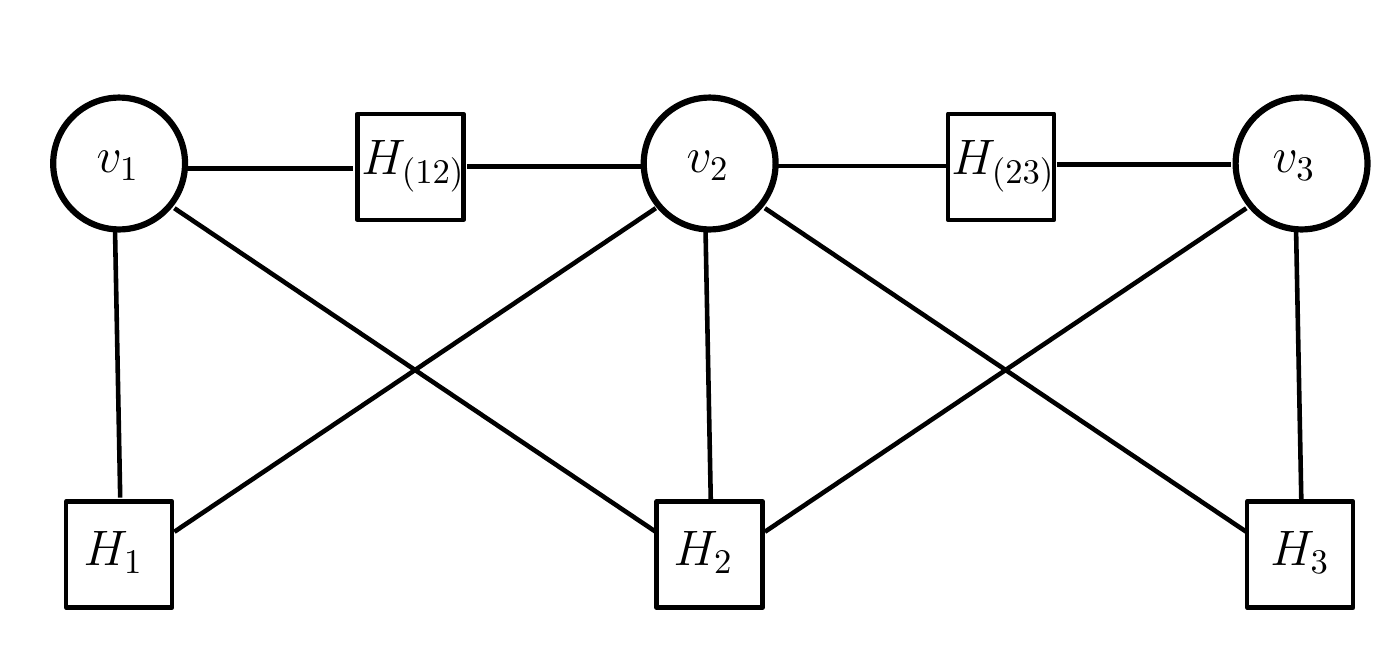}}
    \\
     \subfloat[]{\includegraphics[trim = 170 400 40 200, clip, width = 0.45
     \textwidth]{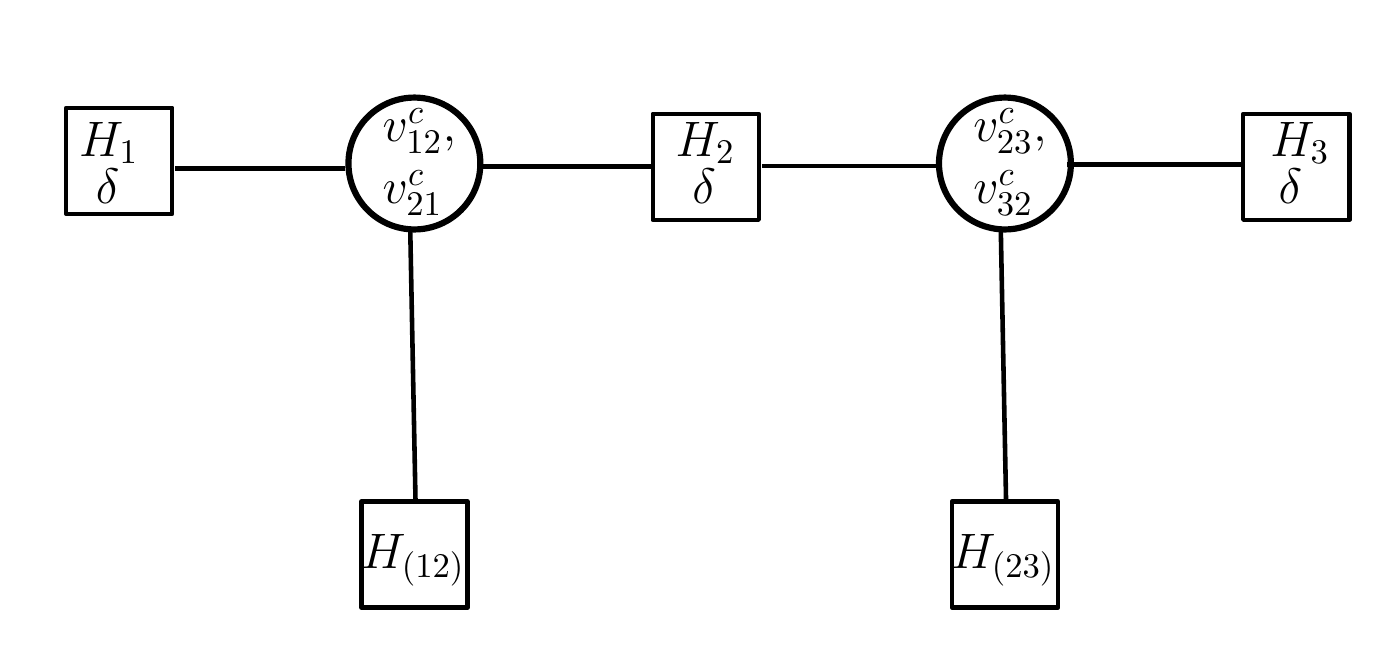}}
    \subfloat[]{\includegraphics[trim = 170 400 40 200, clip, width = 0.45 \textwidth]{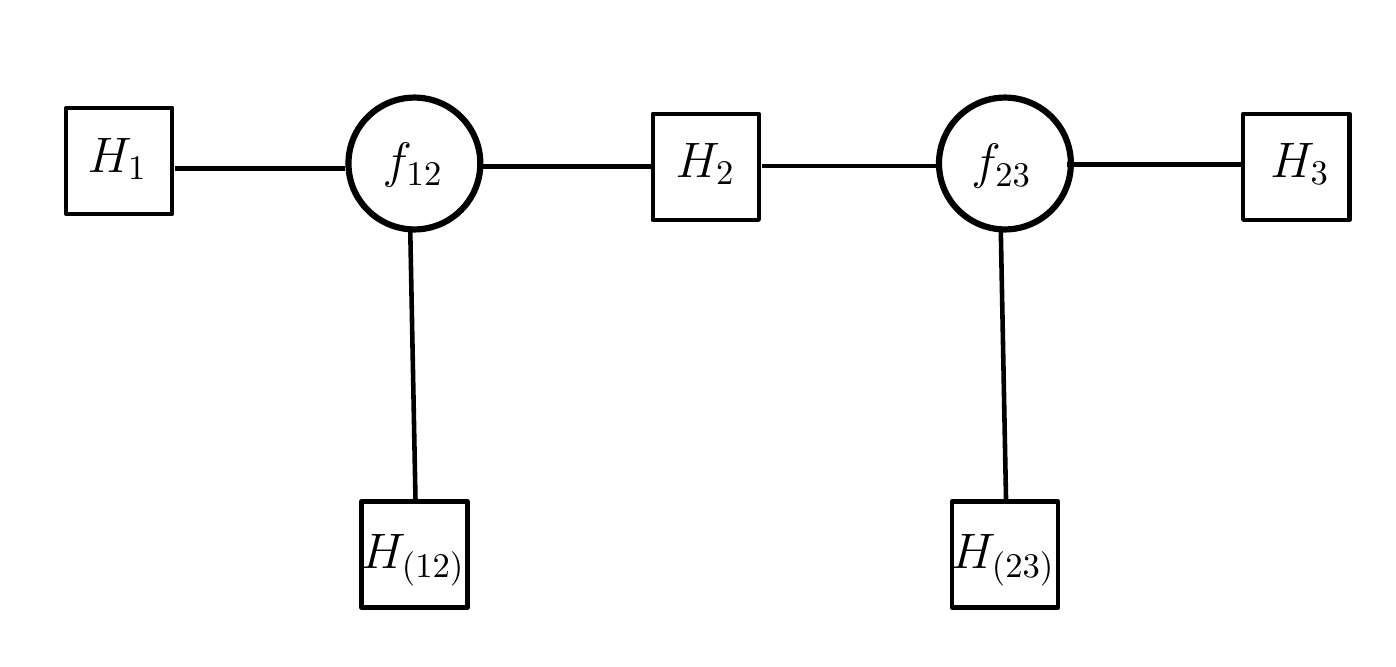}}
    \caption{(a) A simple network, which gives the probability distribution of Eq.~\ref{eq: example_P}. (b) The factor graph assigned in the straightforward way (Section~\ref{sec: BP_basics}) to Eq.~\ref{eq: example_P}.
    (c)~The factor graph assigned to the same distribution by our clustering procedure.
    (d)~The factor graph assigned to the flow-only approximation of the same distribution, as described in Section~\ref{sec: flow-only}.}
    \label{fig: simple_network_2}
\end{figure*}

We defined three different ways of constructing a factor graph from a probability distribution of the form of Eq.~\ref{eq: basicP}:
\begin{itemize}
    \item the naively assigned factor graph (Section~2.1), from here on denoted as $F_v$,
    \item the factor graph clustered according to our procedure of Section~\ref{sec: clustered_graph}, denoted as $F_c$,
    \item the flow-only factor graph (Section~\ref{sec: flow-only}), completely ignoring the vertex variables (according to Eq.~\ref{eq: Pflow}), denoted as  $F_f$.
\end{itemize}
Fig.~\ref{fig: simple_network_2} compares  how these factor graphs look like concretely for the simple network of Fig.~\ref{fig: simple_network}.
$F_c$ and $F_f$ have the same amount of loops as the original network, in this case none.
 $F_v$ has an increased number of loops, where the difference in the amount of loops is given by $2 \cdot \sharp \text{links}$, as argued before in Section~2.1.
These factor graphs lead to three different sets of BP equations (following Eqs.~\ref{eq: BP1}-\ref{eq: BP4}).

\section{A concrete implementation: Bayesian inference for state estimation in power grids}\label{sec:3}
The previous section contains the general formulation of our clustering method; it gives an improved BP for supply networks by considerably reducing the number of loops in the factor graph. In this section we will use a realistic implementation for state estimation in power grids as a concrete test case to show the improvement over the naive assignment of factor graphs. The goal in state estimation problems is to reliably retrieve the underlying state of the system in terms of its variables. The variables are correlated by power flow equations and in principle accessible to measurements, but these are affected by errors, therefore some care is needed to reliably estimate the state.

In AC-power grids, the power flow equations restrict the measured values for active (reactive) power injections 
at vertex $i$, the active (reactive) flows
between vertex $i$ and vertex $j$, as well as the voltages,
given the conductances  and susceptances  of the transmission lines. The DC-approximation to the AC-equations, which we consider in more detail, corresponds to a linearization of the AC-equations which is justified for high-voltage grids when angle differences are small and Ohmic power losses can be neglected. In this case, the DC-approximated AC-equations read
\begin{equation}\label{eq:powerflow}
f_{ij}=B_{ij}(\theta_i-\theta_j)\;,\qquad g_i= \sum_{j\in N(i)}f_{ij}
\end{equation}
with $\theta_i$ the phase angles at the vertices $i$, $g_i$ representing active power injections at vertices $i$, $f_{ij}$ the active power flow. The susceptances $B_{ij}$ are provided in the data sets of the considered grids and $N(i)$ denotes the set of vertices directly connected to vertex $i$ by a transmission line.
In \cite{ritmeester_state_2020} we ignored the first of Eqs.~\ref{eq:powerflow} and considered only the flows as variables characterizing the state  of the system.
This corresponds to the flow-only approximation discussed in Section~\ref{sec: flow-only}.
In principle, the injections $g_i$, the flows $f_{ij}$ and more recently also the angles (via phasor measurement units (PMUs)) are accessible to direct measurements.
However, one can do better than taking these direct measurements for the state estimation and use Bayes' theorem in the form of
\begin{equation} \label{eq: dist_factorization}
    P(\bm{x}|\bm{z}) = \frac{P(\bm{z}|\bm{x}) P_{\text{pr}}(\bm{x})}{P_{\text{pr}}(\bm{z})} \,,
\end{equation}
where $P(\bm{z}|\bm{x})$ is the probability that a state $\bm{x}$ would give data $\bm{z}$ and $P_{\text{pr}}(\bm{x})$ is the prior belief that the state is $\bm{x}$. If no prior knowledge exists about $\bm{x}$, it will be chosen as a uniform distribution. The prior belief over data, $P_{\text{pr}}(\bm{z})$, is independent of $\bm{x}$; hence it only provides a normalization constant and is irrelevant for our purposes.
Bayes' theorem is then used for state estimation and real-time processing of  measurement data.

The injections $\{g_i\}$ and flows $\{f_{ij}\} = \{-f_{ji}\}$  are restricted by the DC-approximated power-flow equations (Eqs.~\ref{eq:powerflow}), so we can write $f_{ij}$ and $g_i$ as functions of the angles, $f_{ij}(\theta_i, \theta_j)$ and $g_i(\theta_i, \theta_j: j \in N(i))$.
Denoting a given measurement as $z_a$ (always a scalar), and the subset of variables which enter the measurement as $\bm{\theta_a}$, we then assume $z_a = f(\bm{\theta}_a) + \xi_a$, where $\xi_a$ are generated independently from Gaussian distributions with known standard deviations $\sigma_a$.
The function $f$ represents the scalar quantity which is measured, specified as a function of the angle variables.
If we consider direct measurements of state variables ($g_i$, $\theta_i$ or $f_{ij}$), then we have $f(\bm{\theta_a})$ equal to  $g_i(\theta_i, \theta_j: j\in N(i))$, $\mathbbm{1}(\theta_i)$ or $f_{ij}(\theta_i, \theta_j)$, respectively.

Measurement errors of this form thus give $P(z_a|\bm{\theta}_a) \sim N(f(\bm{\theta_a}), \sigma_a)$,
which contribute to the joint probability distribution  $P(\bm{z}|\bm{\theta}) = \prod_a P(z_a|\bm{\theta}_a)$.
If no direct measurement is available, it is convenient to write this as a measurement with $\sigma_a \rightarrow \infty$ (and $z_a$ arbitrary).
With Bayes' theorem we get:
\begin{eqnarray} \label{eq: basicP1}
    P(\bm{\theta}\vert \bm{z}_g, \bm{z_}f, \bm{z_}\theta) &=& P(\bm{z}_g, \bm{z_}f, \bm{z_}\theta\vert \bm{\theta}) P_{\text{pr}}(\bm{\theta})\\ \nonumber
    &=& \Big[ \prod_{i} P(z_{g_i}\vert g_i(\theta_i, \theta_j: j \in N(i)) \times P(z_{\theta_i}\vert \theta_i)  
    \Big] \\ \nonumber
    &\times &\Big[ \prod_{(i,j)} P(z_{f_{ij}}\vert f_{ij}(\theta_i, \theta_j))
    \Big] \times P_\text{pr}(\bm{\theta})\,,
\end{eqnarray}
where $\bm{z}_g, \bm{z_}f$ and $\bm{z_}\theta$  denote the set of power injection, flow and angle measurements, respectively.
Thus, this distribution gives the likelihood that the true state is $\bm{\theta}$, given the measurements in $ \bm{z}_g, \bm{z_}f, \bm{z_}\theta$.
In view of the state estimation problem we are interested in calculating marginals of Eq.~\ref{eq: basicP1} such as $P_{(ij)}(\theta_i, \theta_j)\equiv\int\prod_{k\not=i,j} d\theta_k P(\bm{\theta}| \bm{z}_g, \bm{z_}f, \bm{z_}\theta)$ to calculate likely values of the flow $f_{ij}$ and the corresponding phase angles (and similar for other quantities). 
To calculate the marginals, we have to deal with a number of integrals of large products over all vertices and transmission lines.

\subsection{BP for power grid state estimation: Performance in terms of speed and accuracy}
The distribution $P(\bm{\theta}| \bm{z}_g, \bm{z_}f, \bm{z_}\theta)$, as a function of $\bm{\theta}$, is
of the form of the general distribution given in Eq.~\ref{eq: basicP} (assuming a uniform prior).
We can thus use BP to solve the state estimation problem and compare the results for the different factor graphs.

In the following, the subscript $x$ shall indicate for which factor graph $F_x$  is evaluated:
$F_v$ (the straightforward factor graph), $F_c$ (the clustered factor graph), and $F_f$ (the flow-only factor graph).
We use BP to solve the state estimation problem on the IEEE-300 benchmark network \cite{IEEEnetworks}.
The IEEE-300 network is a realistic and heterogenous benchmark network with 300 vertices, 411 links and 112  loops, it is described in more detail in \ref{app: benchmark_networks}. As explained in Sections~\ref{sec: introduction} and \ref{sec: clustering}, the use of the new algorithm refers to the avoidance of extra loops in the associated factor graph. We do not modify  the 112 loops (by clustering some vertices) in the IEEE-grid, but keep them and compare the performance with and without the loops in the assigned factor graphs.
Note that the number of additional loops in the naively assigned factor graph would be $\sum_i\vert N(i)\vert=2\cdot\sharp \; \mbox{links}=2\times 411=822$ on top of the 112 loops of the IEEE-300 grid, and all these additional loops would be short.

We will consider measurements of the power flows, measurements of the power injections, and, if PMUs are assumed to be present, measurements of the phase angles. We discuss two situations: one where all of the variables are measured (measurement devices at every vertex and transmission line), and one --the more realistic case-- where only the flows and injections are measured (i.e., without any PMUs).
The flow and injection measurements are assumed to have an error $\xi_a$ with variance $\sigma^2_a = 10^{-3}$, while the angle measurements, if present, are assumed to have an error $\xi_a$ with variance $10^{-6}$.
Using these values, we randomly draw the measurements $\bm{z}$ following the description in the previous section,
and use BP to find estimates of the state variables by calculating marginals of $P(\bm{\theta}|\bm{z})$.

For comparison we first make use of a  ``damping'' method proposed in \cite{cosovic_state_2016} to improve the convergence of BP on $F_v$ (since numerical simulations have shown that naively running BP on $F_v$ gives diverging estimates).
According to this damping procedure of \cite{cosovic_state_2016}, for each Gaussian message $m_{x \rightarrow y}^t \sim N(\mu_{x\rightarrow y}^t, (\sigma^2)_{x \rightarrow y}^t)$ of Eqs.~\ref{eq: BP1} and \ref{eq: BP2}
one chooses with probability $1/2$ either $\delta=0$ or $\delta=1$ ($P(\delta = 0) = P(\delta = 1) = 1/2$), and updates:
\begin{equation} \label{eq: damping}
    \mu_{x\rightarrow y}^{t + 1} = \delta \times \hat{\mu}_{x\rightarrow y}^{t + 1} + (1-\delta) \times 1/2 \times \Big(\hat{\mu}_{x\rightarrow y}^{t + 1} + \mu_{x\rightarrow y}^{t} \Big),
\end{equation}
where $\hat{\mu}^{t+1}_{x\rightarrow y}$ is the mean of the message that would have been calculated at step $t+1$ without damping. The variance $\sigma^2_{x \rightarrow y}$ is damped equivalently. Thus, with probability $1/2$ the message is updated as usual, and otherwise damped by a factor of $1/2$.
Testing the method according to Eq.~\ref{eq: damping} for different damping parameters and comparing it to the damping algorithm proposed in \cite{pretti_message-passing_2005}, we indeed find that damping according to \cite{cosovic_state_2016} (Eq.~\ref{eq: damping}) improves the convergence the most.
For the results for $F_v$ we will use this ``damped'' version of BP and use it as the best existing alternative, which is still outperformed by our method. For $F_c$ and $F_f$, the algorithm converges without problem also without damping, so we will consider their undamped versions.

\subsection{Results for the factor graphs $F_v$, $F_c$ and $F_f$}
We present our results for the predictions from the three factor graphs $F_v$, $F_c$ and $F_f$. (Details on the implementation are given in \ref{sec: pseudocode}.)
We focus here on the estimation of the flows $\{f_{ij}\}$ (in $F_c$ and $F_v$ these can be retrieved as $B_{ij}(\theta_i - \theta_j)$).
\setcounter{footnote}{0}
BP on $F_v$, $F_c$ and $F_f$ will produce different estimates of the marginals $\{P_{(ij)}(f_{ij})\}$,
which we will denote by $\{b_i^v(f_{ij})\}$, $\{b_i^c(f_{ij})\}$ and $\{b_i^f(f_{ij})\}$, respectively.
To retrieve their accuracy, we need a way to compare them with the `true' marginals $P_{(ij)}(f_{ij})$.

Note that the variables of the factor graph $F_c$ are  tuples $\theta_{ij}^c,\theta_{ji}^c$, the beliefs resulting from $F_c$ depend on $\theta_i,\theta_j$, so that averages or variances of the flows $B_{ij}(\theta_i-\theta_j)$ can be directly calculated using these beliefs as the probability distribution $P_{(ij)}(\theta_i,\theta_j)$ according to Eq.\ref{eq: BP3}, when calculating expectation values. In contrast, the variables of $F_v$ are $\theta_i$,  the resulting beliefs depend on $\theta_i$ separately. In this case the variance of the flow  $f_{ij}$ cannot simply be obtained as the sum ($B_{ij}^2$ times the variances of $\theta_i$ and $\theta_j$), since $\theta_i$ and $\theta_j$ are correlated. The marginal distribution of $f_{ij} = B_{ij} (\theta_i - \theta_j)$, in particular its variance, must be calculated from Eq.\ref{eq: BP4}.

\begin{figure}
    \centering
    \subfloat[]{\hspace{-1cm}\includegraphics[width = 0.8 \textwidth]{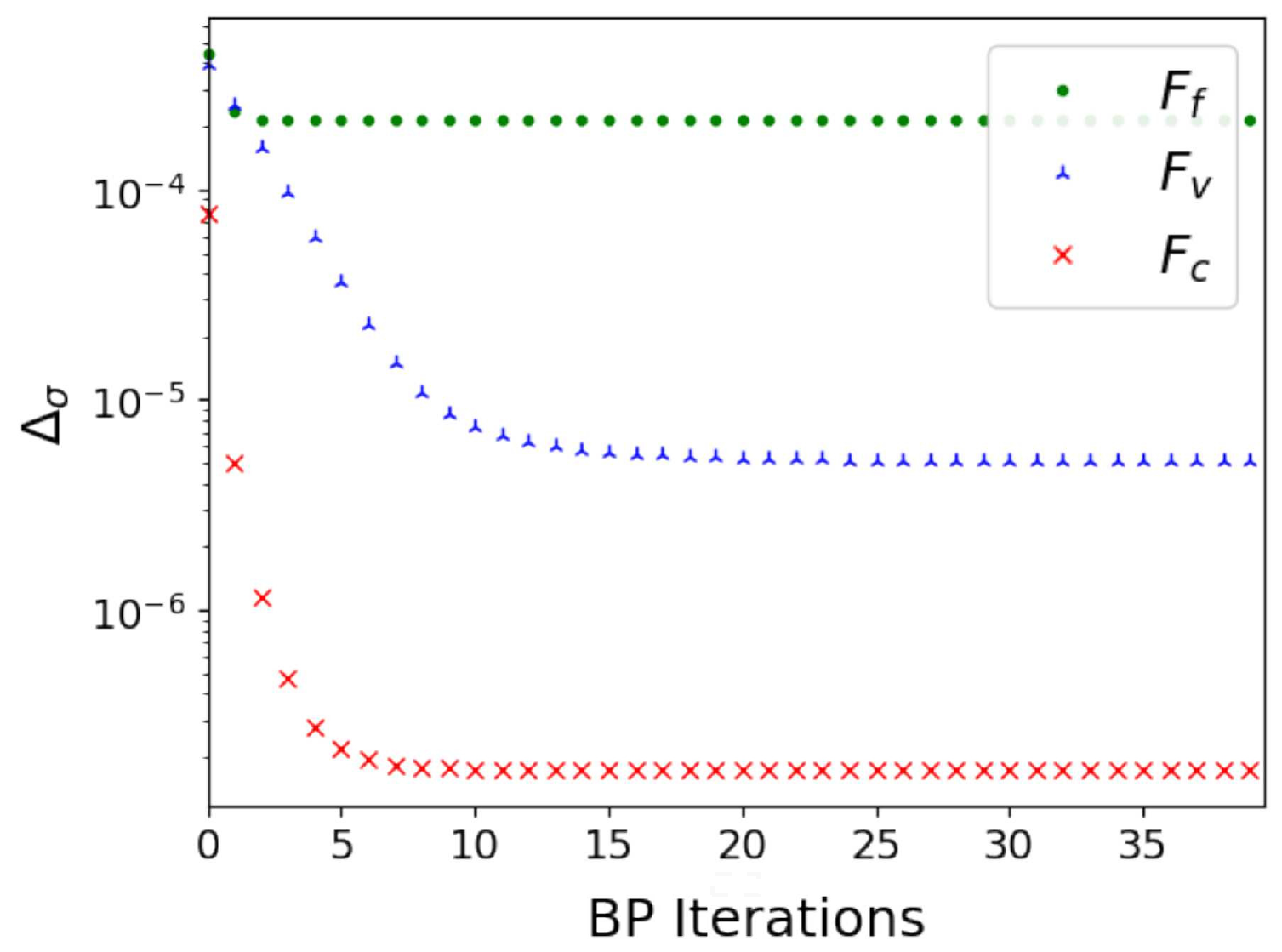}}\\
    \subfloat[]{\hspace{-1cm} \includegraphics[width = 0.8 \textwidth]{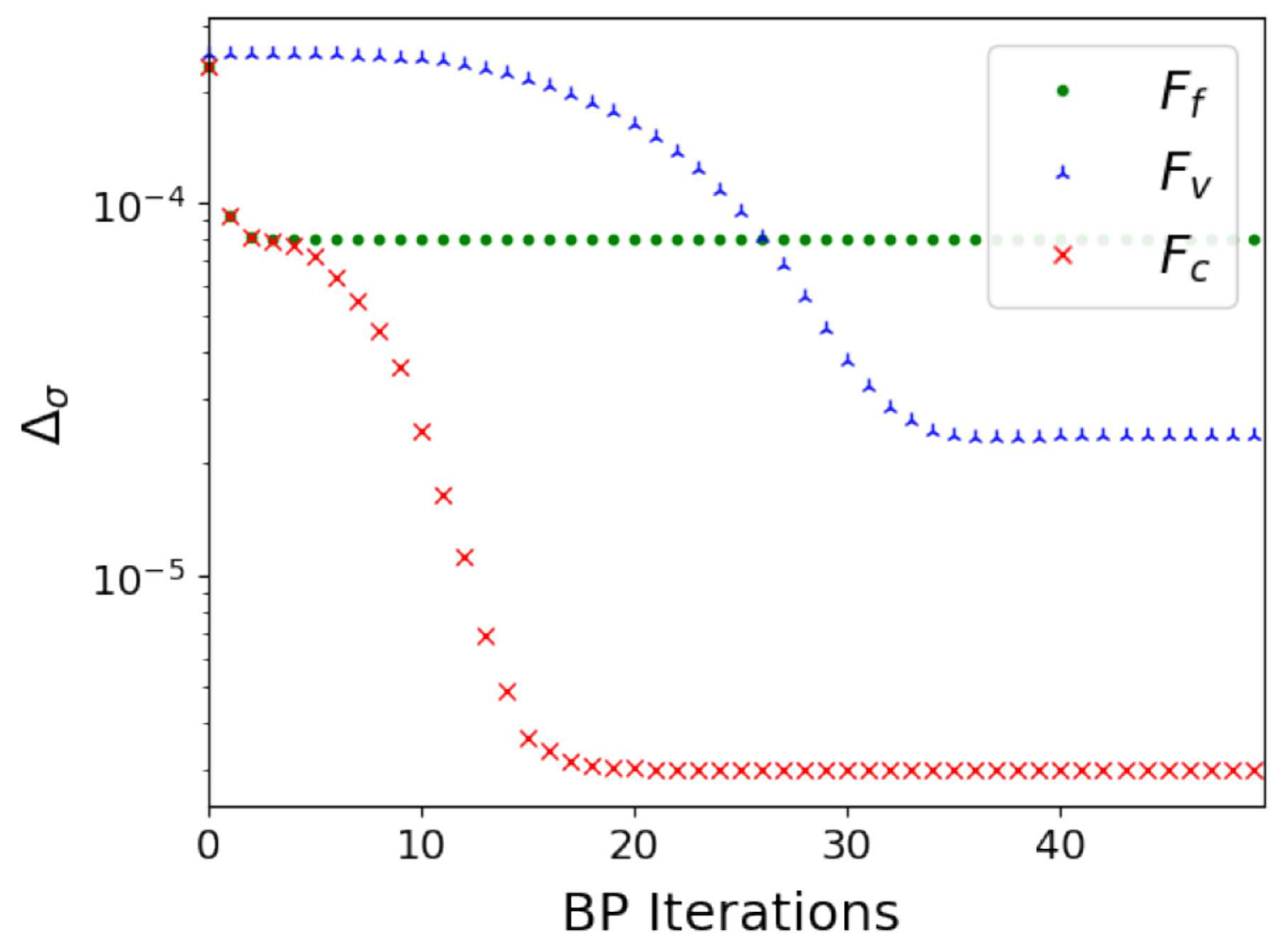}} \\
    \caption{The saturation of the variance predicted by BP on the different factor graphs, as measured by the average square error $\Delta_\sigma$ (see Eq. \ref{eq: error_sig}), showing that the clustered version of BP is most accurate: (a) with angle measurements, (b) without angle measurements. The variances predicted by $F_c$ and $F_f$ do not depend on the values of the chosen measurements. The variances predicted by $F_v$ are slightly different every time BP is run because of the probabilistic damping (Eq.~\ref{eq: damping}); here an average over 100 random measurement sets is shown. Note that $F_v$ corresponds to the standard factor graph assignment and serves as the best version of existing alternatives.}
    \label{fig: var_convergence}
\end{figure}

\begin{figure}
    \centering
   \subfloat[]{ \hspace{-1cm}\includegraphics[width = 0.8 \textwidth]{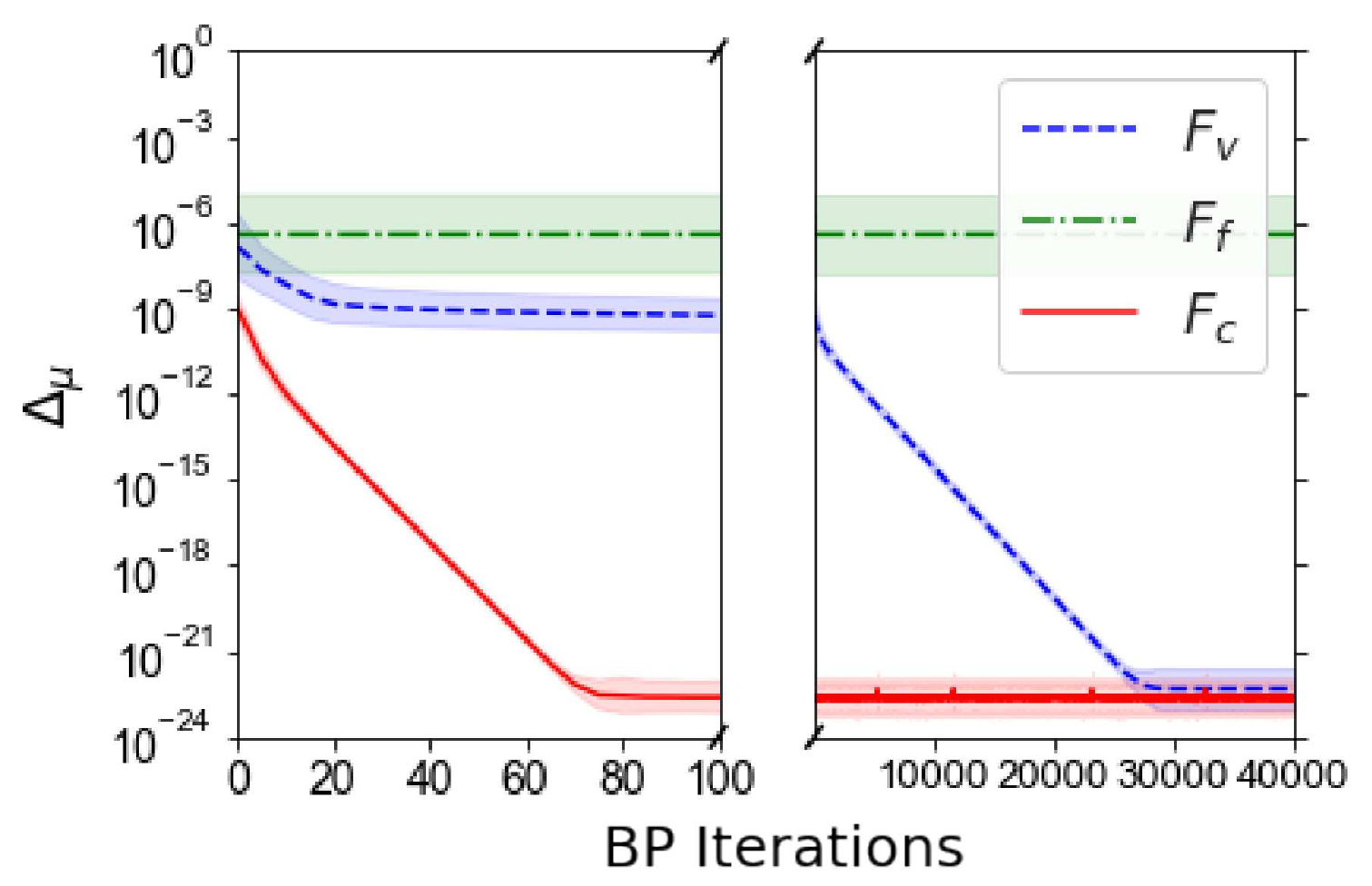}} \\
   \subfloat[]{ \hspace{-1cm}\includegraphics[width = 0.8 \textwidth]{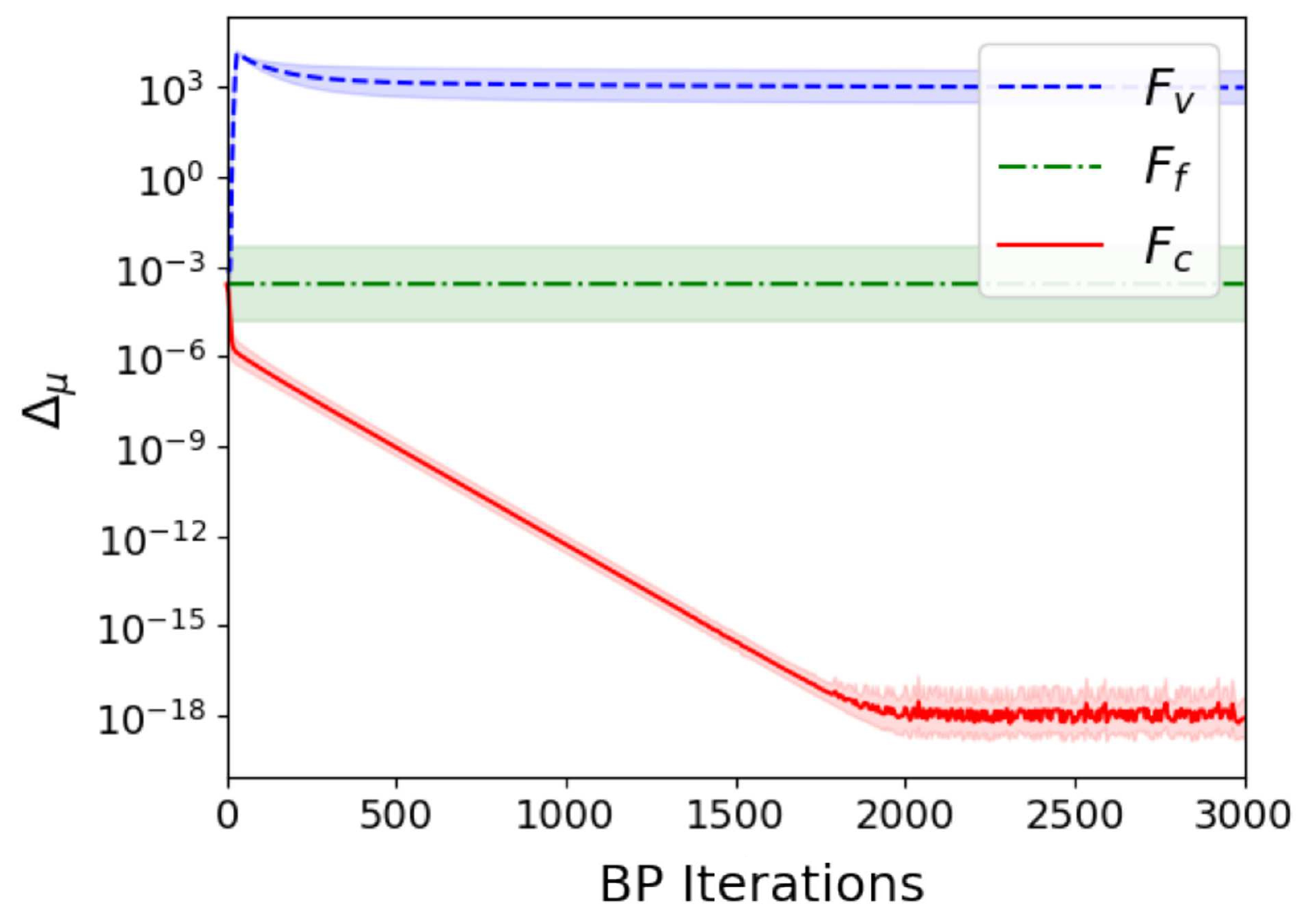}} \\
    \caption{The convergence of the mean predicted by BP on the different factor graphs, as measured by the average square error $\Delta_\mu$ (Eq.~\ref{eq: error_mu}), as a function of the number of BP iterations, showing that BP with only flows ($F_f$) is fastest but not very accurate, while clustered BP ($F_c$) converges reasonably fast and gives the exact answer (up to machine precision).
    When un-clustered BP ($F_v$) \textit{does} converge it also gives the exact answer; this takes, however, very long: (a) with angle measurements, (b) without angle measurements. The predictions depend on the values of the measurements; here the lines give an average of $\Delta^\mu$ over $100$ sets of random measurements. The filled region gives the standard deviation of $\Delta^\mu$ between different random sets of measurements.}
    \label{fig: mean_convergence}
\end{figure}

Since we assume all factors are Gaussian, the marginal distributions of the flows are Gaussian as well, so we denote them by their mean and standard deviation as $P_{(ij)}(f_{(ij)}) \sim N(\mu_{(ij)}, \sigma_{(ij)})$. Here the subscript $(ij)$ represents the flow variables in the following. We calculate the means $\{\mu_{(ij)}\}$  via the least-squares approach and the standard deviations $\{\sigma_{(ij)}\}$ by a matrix inversion.  Both the means and standard deviations that are calculated in this way are presumed to set the accurate benchmark for a comparison to the accuracy of the different implementations of BP.
Each implementation of the factor graph with $x\in\{v,f,c\}$  gives an estimate $b_{f_{(ij)}}^x(f_{(ij)}) \sim N(\mu_{(ij)}^x, \sigma_{(ij)}^x)$, which should be close to the benchmark values $\mu_{(ij)}, \sigma_{(ij)}$. For a chosen method $x$ we  summarize the estimates into the average square error of $\{\mu_{(ij)}^x\}$ and $\{\sigma_{(ij)}^x\}$ by defining:
\begin{eqnarray}
    \Delta_\mu &\equiv& \frac{1}{411}\sum_{(ij)}(\mu_{(ij)}^x - \mu_{(ij)})^2 \label{eq: error_mu}\\
    \Delta_\sigma & \equiv& \frac{1}{411}\sum_{(ij)}(\sigma_{(ij)}^x - \sigma_{(ij)})^2 \label{eq: error_sig}\,,
\end{eqnarray}
where the sum runs over all $411$ links of the IEEE-300 network.
We use these deviations to assess the accuracy of the estimates provided by the different implementations of BP. Since random generation of the measurement data $\bm{z}$ leads to differing errors, we repeat the procedure $100$ times. For a fair comparison, we note that per BP iteration the wall-clock time for a native Python 3.7 implementation (on an Intel i5-2400 processor) of $F_f$, $F_c$ and $F_v$ takes  $0.01$ s, $0.03$ s, and $0.1$ s, respectively. The implementation is given in the supplementary material. Native Python is relatively very slow (up to two orders of magnitude slower than other faster implementations) and the calculations can be massively parallelized, so these time-scales can be significantly reduced (the relative speed of BP on the different factor graphs are expected to remain more or less unchanged).\\
Fig.~\ref{fig: var_convergence} shows how the error on the variance $\Delta_\sigma$ saturates for BP on the different factor graphs.
For all situations discussed above, the estimated variances converge fast to small values. The estimates provided by $F_c$ are significantly more accurate than those provided by $F_v$, which are again significantly more accurate than those provided by $F_f$. Note that $F_v$ corresponds to the standard implementation of BP with a naive factor graph assignment.
\vskip1pt Focusing on the estimates for the mean, the convergence of the estimates for the different scenarios are shown in Fig.~\ref{fig: mean_convergence}. Fig.~\ref{fig: mean_convergence}(a) shows the situation where PMU measurements are included.
After convergence, the means predicted by $F_v$ and $F_c$ are both exact (as is in general true for means predicted by Gaussian BP \cite{weiss_correctness_2001,bickson_gaussian_2009}). However, $F_c$ converges in much less iterations than $F_v$, by around a factor of 400.
Looking at the mean estimates for the situation without PMUs,
as shown in Fig.~\ref{fig: mean_convergence} (b), the situation is similar.
$F_c$ converges quickly to the exact answer.
Although it is not shown here, experiments indicate that eventually the mean predicted by $F_v$ does converge.
However, the time scale over which it converges is so much larger (about $10^6$ iterations) that it renders the final estimate practically irrelevant. In practice, in the absence of angle measurements even $F_f$ performs better than $F_v$.

In summary of Figs.~\ref{fig: var_convergence} and \ref{fig: mean_convergence}, it should be emphasized that our comparison refers to the performance of different BP-algorithms, differing by the assigned factor graph, for which the newly proposed assignment $F_c$ performs best. In contrast to other approaches such as least-squares or quasi-Newton methods, the general supremacy of  BP based algorithms was demonstrated already in \cite{cosovic_state_2016, cosovic_observability_2020, ritmeester_state_2020}, as mentioned in the introduction.

\section{Outlook to other applications of the algorithm to power grids}\label{sec:5}
\newcommand{\approptoinn}[2]{\mathrel{\vcenter{
  \offinterlineskip\halign{\hfil$##$\cr
    #1\propto\cr\noalign{\kern2pt}#1\sim\cr\noalign{\kern-2pt}}}}}
\newcommand{\appropto}{\mathpalette\approptoinn\relax}
Our clustering rule for loopy factor graphs can be used for any distribution of the form of Eq.~\ref{eq: basicP}.
It thus applies for BP to a variety of supply networks,
if the flows are conserved at the vertices and are determined by variables at the vertices ($\bm{v}$, in our notation).
In the following we mention different versions of power flow in electricity grids and discuss applications to gas-pipeline networks and fluid flow networks in the appendix.
Problems that are studied in relation to power flow include power flow analysis, i.e., solving the power flow equations, similar to what we analyze for the gas pipe network in \ref{sec:4}, optimal power flow \cite{dvijotham_graphical_2017}, state estimation \cite{cosovic_state_2016, cosovic_distributed_2019}, as considered  in Section 3, and optimization under uncertainty \cite{summers_stochastic_2014, bienstock_chance-constrained_2014}.
In these applications, typically three different power flow equations are distinguished:
\begin{itemize}
 \item The DC-approximation as used in Section~\ref{sec:3} as an approximation of the AC-equations, valid for high voltages at low power losses.
        Note that despite the name, the application of the DC-approximation is to AC-networks.
    \item In a Direct Current (DC)-network, the size of the current $I_{ij}$ between vertices $i$ and $j$ is given by Ohm's law $I_{ij} = (V_i - V_j)/R_{ij}$, where $V_i$ and $V_j$ are the voltages at vertices $i$ and $j$ and $R_{ij}$ is the resistance of the link connecting vertices $i$ and $j$.
    DC is used in low-voltage distribution grids and in very long distance transmission \cite{machowski_power_2020}.
    \item Alternating Current (AC) is typically used for high voltage-long distance transmission \cite{machowski_power_2020}. The network voltages $\{V_i(t)\}$ oscillate at a constant frequency $\omega$, such that $V_i(t) = \sqrt{2}|V_i|\sin(\omega t + \theta_i)$.
   Here $|V_i|$ is  the voltage magnitude. Together with the phase angles $\theta_i$, they can be determined from the equations
  \begin{eqnarray}\label{eq20}
        f^P_{ij} = |V_i| |V_j|\big[G_{ij} \mbox{cos}(\theta_i - \theta_j) + B_{ij}\mbox{sin}(\theta_i - \theta_j)\big] \nonumber\\ -|V_i|^2G_{ij}\,,   \\ \nonumber
        f^Q_{ij} = |V_i| |V_j|\big[G_{ij} \mbox{sin}(\theta_i - \theta_j) - B_{ij}\mbox{cos}(\theta_i - \theta_j)\big] \\ +  |V_i|^2B_{ij},
   \end{eqnarray}
    where $f^P_{ij}$ and $f^Q_{ij}$ are  the active and reactive power flow, each of which is conserved at the vertices.
    Solving the equations requires the specification of the quantities $B_{ij}$ and $G_{ij}$ for each transmission line, known as the susceptance and the conductance, respectively, which can be calculated from the impedance and resistance of the transmission line \cite{machowski_power_2020}.
    To implement AC in our framework, the tuples ($\theta_i$, $|V_i|$) should be considered as the vertex variables $v_i$.
    Due to the nonlinearity of Eqs.~\ref{eq20}, BP should be combined, for example, with GN, as in \cite{cosovic_distributed_2019}.
\end{itemize}

For power grids, a situation with uncertain costs due to fluctuations in uncertain power injections by renewable resources was investigated in \cite{summers_stochastic_2014, bienstock_chance-constrained_2014}.
In this case each possible production assignment leads to a different distribution of the form of Eq.~\ref{eq: basicP}.
Uncertain power injections may enter the probability distribution $P(\bm{v})$ of Eq.~\ref{eq: basicP}
as a product over all vertices $\prod_i\exp{\big\{-[g_i-\sum_{j\in N(i)}f_{ij}(v_i,v_j)]^2/(2\sigma_i^2)\big\} }$ if the production fluctuates according to such a Gaussian distribution with variance $\sigma_i^2$ around some mean injection $g_i$,
while fixed and controllable production at vertices $k$ would multiply this term by $\prod_k\delta(g_k-\sum_{j \in N(k)}f_{kj}(v_k,v_j))$.
$P(\bm{v})$ induces a distribution of flows $P(\{f_{ij}\})$ which directly indicates possible overflows of transmission lines.  BP may then be used to calculate average costs induced by the production $g_k$ at the set of controllable production vertices.
These costs can furthermore include probabilistic constraints on $\{g_k\}$ which enforce, for example, that severe link overloads are rare \cite{summers_stochastic_2014}.
In \cite{altarelli_stochastic_2011} it is shown that such an optimization can be performed with BP by using the fact that marginals of the distribution satisfy the BP equations (to be given below).
This is mathematically similar to the 'survey propagation' studied in \cite{del_ferraro_cavity_2014} and investigation of large deviations given in \cite{rivoire_cavity_2005}.

\section{Conclusions}\label{sec:6}
For applications of BP we considered a new assignment method of factor graphs which avoids the generation of additional loops as compared to the original supply networks. Our method applies to state estimation or optimization problems whose state can  be described by a probability distribution that factorizes over the vertices and over the links of the network. If these distributions are summed or integrated over upon marginalization, BP provides an efficient way of organizing the sum or integrals over these products.
In the naive assignment of a factor graph, the variables on the original supply network are chosen as variable nodes on the factor graph, while the factors of the probability distribution determine the factor nodes on the factor graph. Constraints and physical laws between the variables on the original grid may then induce additional loops on the factor graph that are detrimental for the convergence speed of BP. Our main goal was to avoid these loops.

This means that our algorithm does not address the handling of loops in the graphical representation of the original supply networks, which may differ in size and reflect the original network architecture. Such loops will survive our factor graph representation and may impede the convergence or accuracy of BP. Our method is supposed to complement other methods such as loop expansions and additional clustering rather than replacing them. For practical applications we have furthermore focussed on cases in which the messages themselves are Gaussian functions so that the messages reduce to a few real numbers to be sent. In particular we have assumed Gaussian distributed errors in the state estimation problems, and used successive Gaussian approximations in the steady state analyses.

In the cases we considered, additional loops in the factor graphs result from constraints between the variables which are analogues to the two Kirchhoff laws in power grids, one corresponding to flow conservation at vertices (the flow in general being electricity, gas, water, air, soil, traffic), the second one restricting a quantity related to energy (voltage, pressure, time, other costs) along loops in the grid. The shared mathematical structure of these constraints explains the wide range of applicability of our algorithm. When the resistance in the transmission lines of the original grid are depending on the flow, the analogue of Kirchhoff' s second law amounts to a nonlinear relation (differently from Ohm's law). In this case an additional  iterative method such as the Gauss-Newton method is required, and BP can be applied in the intermediate steps.

While our algorithm enhances the accuracy and improves the convergence speed, the additional computational effort is moderate.
We have explicitly worked out this approach for the state estimation on a benchmark power grid. In the appendix we describe the state determination on two benchmark gas-networks with nonlinear flow relations, the latter case detailed in \ref{sec: clustered_graph}. We compared the performance of three factor graphs: the naively assigned, the newly proposed clustered factor graph, and the flow-only factor graph. The naively assigned one can lead to accurate results, but at the price of slow convergence if at all. The flow-only factor graph ignores constraints from (the analogue of) Kirchhoff's second law, it is thus less accurate but useful for a first estimate and fast. The clustered factor graph is both fast and accurate and -combined with an iterative procedure in case of nonlinear constraints- it is widely applicable.
Further applications to other indicated supply networks should be worked out in the future.
\section*{Acknowledgment}
We thank the Bundesministerium f\"ur Bildung und Forschung (BMBF) (grant number 03EK3055D) for financial support.\\

\section*{Author contribution statement}
T. R. proposed and implemented the algorithm and performed the simulations. Both authors discussed and contributed to the manuscript. H.M.-O. is responsible for its final form.

\appendix

\section{Steady state analysis in natural gas-pipeline networks}\label{sec:4}
Why is a steady state analysis important? Gas transmission network operators are obliged to offer as much freely allocable capacity as possible. They are supposed to ensure that gas traders or consumers feed in or withdraw gas at their entries or exits without being concerned about the impact on the overall grid. On the other hand, to meet the statutory carbon emission targets, the use of natural gas has to decline in the course of time. New sources of alternatives (hydrogen, biogas) are discussed so that the distribution network may be challenged by a mixture of gas input. In this case, a steady-state  analysis of a gas network is required under varying operational conditions. In steady-state analyses, nodal pressures and pipe flows are computed for given values of source node pressures and gas consumption.

We choose our benchmark gas-networks from GasLib \cite{schmidt_gasliblibrary_2017}. GasLib is a collection of technical gas network descriptions as well as ``contract based nomination data''. It is based on real-world network data from the gas transport company Open Grid Europe GmbH. ``The data are distorted in order to yield a realistic gas network that is significantly different from the original'', but to serve as benchmarks for test simulations.

In analogy to state estimation problems as we considered for power grids, state estimation or leak detection may be applied to  gas (or water) transport systems to convert system measurements into reliable information on the network state. In our application here,  instead we want to determine the steady state, that is the gas flow and pressures throughout a gas-pipe network, based on input that is assumed to be accurate \cite{abeysekera_steady_2016, walski_history_2006, coelho_considerations_2007, brikic_gas_2011}. We will perform these calculations with BP for the two realistic natural gas networks (Gaslib-134 and Gaslib-40), 
in order to compare the factor graphs described in Section~\ref{sec:2}.
For the gas-pipe networks, the data of the benchmark networks allow in principle a complete determination of flows and pressures at all vertices of the network. However, to illustrate our method, we do not make use of the input values of the three largest  generators of the gaslib-134, but specify instead the pressures at these stations, so that the unknown variables (flow, pressures, injections) are in principle fully determined, but less easily accessible. This mimics a realistic situation in a gas network and serves as test bed for our method.
To ensure that the system is fully determined, at each vertex either the injection of gas (positive or negative) or the pressure has to be specified.
Given a gas flow equation relating flow $Q_{ij}$ through a given pipe $(ij)$ with the pressure $p_i$ and $p_j$ at the ends, the pressure at the vertices and the flows through the pipes can then be calculated.

As a gas-flow equation we  use
\begin{equation}\label{eq15}
Q_{ij} = a_{ij}\mbox{sgn}(p_i^2 - p_j^2)\big(p_i^2 - p_j^2\big)^{0.5}
\end{equation}
for coefficients $a_{ij}$ specific to each transmission line. For more details we refer to Section~5.
Combined with the continuity equation
\begin{equation}\label{eq16}
g_i = \sum_{j \in N(i)} Q_{ij},
\end{equation}
where $g_i$ is the injection at vertex $i$,
the pressure and flows throughout the network can then be calculated. Treating the flow equations
as constraints on $\{p_i\}$ and following Section~2, the equations can be solved by BP. Eq.~\ref{eq16} (Eq.~\ref{eq15}) corresponds to the first (second) Kirchhoff's law, respectively.

\subsection{Steady state analysis for gas networks: BP combined with Gauss-Newton}

To explicitly show the application of BP to this problem, we first map the solution to the equations to a (least-squares) minimization problem, a standard method in the numerical solution of non-linear equations.
(As for the power state estimation, finding the means (and more generally the Maximum Likelihood Estimate \cite{cosovic_distributed_2019}) can be formulated as a least-squares minimization problem, which we used as benchmark for comparison.)
For convenience we use the square of the pressures as variables, rather than the pressures themselves and define $v_i \equiv p_i^2$. 
We denote the set of known (squared-) pressures $\bar{v_i}$ by $\bar{\bm{v}}$, the set of known injections $\bar{g}_i$ by $\bar{\bm{g}}$, and assume that these quantities fully determine the system.
We then define the cost function in terms of $v_i,v_j$: \vspace{-0.3cm}
\begin{equation} \label{eq: cost_function}
    C(\bm{v}) \equiv \sum_{\bar{g}_i \in \bar{\bm{g}}} (\bar{g}_i - \sum_{j \in N(i)} Q_{ij}(v_i, v_j))^2/2
             + \sum_{v_i \in \bar{\bm{v}}} (\bar{v}_i - v_i)^2/2,
\end{equation}
with $Q_{ij}(v_i,v_j)=a_{ij} \mbox{sgn}(v_i-v_j)(v_i-v_j)^{0.5}$ from Eq.~\ref{eq15}.
One can verify that the global minima correspond to sets of pressures such that $C(\bm{v}) = 0$ and such that the equations are satisfied.
Using Eq.~\ref{eq: optimization_as_P} to turn the minimization into a probability distribution gives a distribution of the form of Eq.~\ref{eq: basicP}, with a factor of type $H_i$ for each vertex.
The factors of type $\{H_{ij}\}$ are absent in this case, but can be included by simply setting the $H_{ij}$ equal to a constant. Thus we can apply our clustering procedure to let BP effectively calculate the solutions of these equations.

In a direct application of BP, the numerical computation of the message update Eqs.~\ref{eq: BP1}-\ref{eq: BP4} is, however, complicated by the non-linearity of the flow equation. This means that the cost function (Eq.~\ref{eq: cost_function}) is non-quadratic.
Consequently the corresponding distribution is not Gaussian, and the messages do not maintain a Gaussian shape.
This would require a more sophisticated method to represent the messages and to perform calculations with such messages (as discussed in Section~\ref{sec: BP_algorithm}).

The difficulty in handling non-linearities is of course not restricted to BP.
Methods such as Newton's method can then be used to make successive approximations of the cost function, eventually converging to a minimum of Eq.~\ref{eq: cost_function}.
If each approximation gives a quadratic cost function, the quadratic minimization can be solved efficiently with BP using only Gaussian messages, thus allowing for a simpler implementation of BP.
For a quadratic cost function it is the mean of the associated distribution (Eq.~\ref{eq: optimization_as_P}) that gives the minimum of the cost function. The mean is furthermore independent of $T$,
so we can conveniently set $T = 1$ rather than taking the limit.\\
Methods that can implement BP in this way are, for example, apart from Newton's method, the Gauss-Newton (GN-)method (employed in \cite{cosovic_distributed_2019} to solve the AC state estimation problem and in \cite{han_enabling_2018} for water network state estimation), or the fully parallelized  BP procedure proposed in \cite{wong_inference_2007}.
To give an explicitly worked out example, here we use a modified Gauss-Newton method, combined with BP to perform the calculations at each iteration of Gauss-Newton.
The procedure is as follows:
\begin{enumerate}
    \item Set an initial guess for the flows, $\{Q_{ij}^\ast \}$, obtained from the flow-only-approximation (ignoring the $v_i$ variables). See \ref{sec: pseudocode} for details on the implementation.
    \item At each iteration:
    \begin{itemize}
        \item Generate a guess $\{{v_i}^\ast \}$ from the guess $\{Q_{ij}^\ast\}$,
        the known values $\{\bar{v}_i\}$ and the flow equation \\ $Q_{ij} = a_{ij}\mbox{sgn}(v_i - v_j)|v_i - v_j|^{0.5}$.
        This can be done in a straightforward way with time complexity linear in the system size (see the supplementary material).
        \item Linearize the flow equation around this guess $\{{v_i}^\ast \}$ by expanding $Q_{ij}(v_i, v_j)$ to first order in $(v_i - v_j)$.
        \item Using this linearization, the cost function (Eq.\ref{eq: cost_function}) becomes quadratic in $\{v_i\}$.
        These equations can then be solved efficiently by Gaussian BP with our clustering method to generate a new guess $\{Q_{ij}^\ast \}$ (see \ref{sec: pseudocode} for details on the implementation).
    \end{itemize}
    \item If the guesses $\{Q_{ij}^\ast\}$ and $\{{v_{i}}^\ast\}$ have converged to a fixed value, then the non-linear equations have been solved. Stop the iteration if a desired tolerance threshold is reached.
\end{enumerate}
The unmodified GN-algorithm would  follow the same steps, but  omit the intermediate flow guesses $\{Q_{ij}^\ast\}$. Newton's method would omit the linearization altogether, and instead expand $C(\bm{v})$ directly to second order in $\bm{v}$ around a guess $\bm{v}^\ast$. Newton's method does not allow an efficient calculation of the BP-messages given in \ref{sec: pseudocode}, and requires instead  a large number of matrix inversions. \\
\\
The modification of the GN-algorithm we use here instead takes advantage of the linearity of flow conservation.
Comparing it to the unmodified GN algorithm we found that it strongly improves the convergence.
On tree networks, the modified GN-algorithm gives the exact result in a single iteration (which is true for neither the unmodified GN-algorithm nor for Newton's method).
For the GasLib-134 network (see the next section),  instead we restrict ourselves artificially  and use only a single $\bar{v}_i$ (instead of all $\bar{\bm{v}}$) when generating a guess $\{v_i^\ast\}$ from $\{Q_{(ij)}^\ast\}$,
to allow a comparison of  $F_c$ and $F_v$ in an iterative setting.

\subsection{Results for GasLib networks as benchmarks}
Here we compare BP on the different factor graphs $F_v$ and $F_c$. Since $F_f$ ignores constraints imposed by the vertex variables (pressures), the flow-only approximation is under-determined and cannot give a valid solution (in contrast to the state estimation problem studied in \ref{sec:4}, where the system is over-determined).
The flow-only approximation is, however, still useful as an initial guess for the GN-procedure (simply setting undetermined variables to $0$), which we have used here.

\begin{figure}
    \centering
    \includegraphics[width = 0.95 \textwidth]{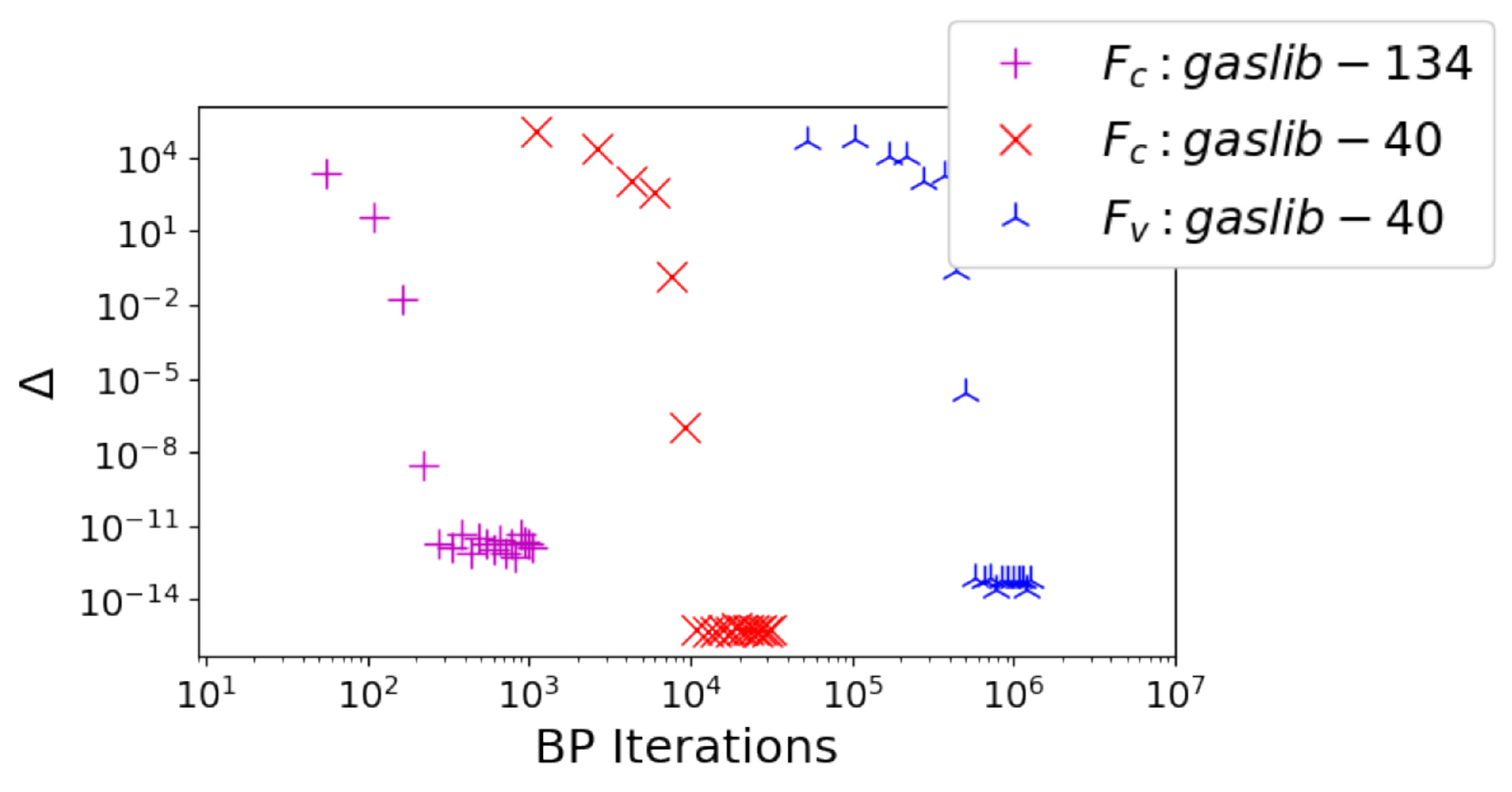}
    \caption{Total BP-iterations needed for convergence on the factor graphs $F_v$ and $F_c$ for GasLib-40 and $F_c$ for GasLib-134,  as measured by the average square error $\Delta$ (Eq.~\ref{eq: square_error_gas}). Each marker indicates a single GN-step. Non-zero $\Delta$ after convergence is due to numerical round-off and finite tolerance. Results for $F_v$ on GasLib-134 are not displayed as here BP does not converge at all.}
    \label{fig: gaslib-GN}
\end{figure}

The GasLib-134 has 134 vertices, $133$ links and no loops, while GasLib-40 has 40 vertices, $45$ links and 6 loops.
We presume that for GasLib-40 the pressure at one vertex is known (\ref{app: benchmark_networks}), while the injection at all other vertices are known. For GasLib-134 we presume that the pressure at three vertices is known (\ref{app: benchmark_networks}), while the injection at all other vertices are known to mimic a realistic situation.
This fully determines the system, such that the flows $Q_{(ij)}$ can be calculated.
For a guess $Q_{(ij)}^x$, given by a method $x \in \{v, c\}$, we compare the guess to the actual flows $Q_{(ij)}$ (obtained by a least-squares procedure) by defining the average square error:
\begin{equation} \label{eq: square_error_gas}
    \Delta \equiv \frac{1}{N}\sum_{(ij)}(Q_{(ij)}^x - Q_{(ij)})^2,
\end{equation}
where $N$ is the number of links of the network used.
We use the GN-procedure, where for each step BP is either run on $F_v$ or on $F_c$.
Thus, the difference  results  only from the way of how each step of the GN-procedure is calculated.
The results are given in Fig.~\ref{fig: gaslib-GN}, where the average square error $\Delta$ (Eq.\ref{eq: square_error_gas}) is shown as a function of the total amount of GN-steps and BP-iterations.
Very few GN-steps are needed for the convergence to the correct solution;
the main difference is in the amount of BP-iterations that is needed to complete a single GN-step.

For each GN-step for the GasLib-40 network, we iterate BP until convergence.
(Specifically, the stopping criterion is as follows: We take an amount of iterations $T$ such that the square difference between the estimate at $T$ and the estimate at $T/2$ is smaller than $10^{-10}/40$ (more lenient stopping criteria may decrease the amount of iterations needed by a factor $\sim 3$).)
Fig~\ref{fig: gaslib-GN} shows that this takes about ~1500 iterations per GN step for BP on $F_c$, compared to ~50000 iterations for each BP step on $F_v$.
On top of this, each BP iteration on $F_v$ takes about three times as much CPU time as an iteration on $F_c$,
such that in terms of speed $F_c$ outperforms $F_v$ by a factor $\sim100$.

For the network GasLib-134 the contrast is even stronger, here BP on $F_c$ and $F_v$ perform according to two extremes:
This network is a tree network, in which the longest path has a length of $55$.
Consequently, the associated factor graph $F_c$ is also a tree graph, and within each GN-step BP on $F_c$ converges to the exact result after $55$ iterations (Fig.~\ref{fig: gaslib-GN}).
This is not true for $F_v$, which has many loops despite GasLib-134 being a tree network. We find that BP on $F_v$ shows oscillatory behaviour and does not appear to converge at all (even after millions of iterations).

\section{Outlook to further applications to fluid flow}
Typical calculations for fluid flow are similar to those studied for natural gas in \ref{sec:4},
where the aim is to calculate the pressures and flows throughout the network, given that we know the injections and pressures at some selected nodes (such that the system is fully determined).
Such calculations are required  for water or gas infrastructure in urban environments, or for long-distance transmission \cite{ abeysekera_steady_2016,walski_history_2006, coelho_considerations_2007, brikic_gas_2011}.
Other calculations in the context of such infrastructure networks refer to optimization \cite{schmidt_gasliblibrary_2017} and state estimation \cite{ahmadian_behrooz_modeling_2015}.
The effect of fluctuating inputs or link damage on the fluctuations of pressure and flow (\cite{katifori_damage_2010}) can be similarly effectively studied with the BP framework constructed in Section~\ref{sec:2}.
For explicit expressions of the relevant equations for fluid flow, three regimes can be distinguished:
\begin{itemize}
    \item Pressurized gases, which we studied with a modified GN-method combined with BP in \ref{sec:4}.  Relations between flow and pressure are partly theoretically and partly phenomenologically motivated \cite{abeysekera_steady_2016, coelho_considerations_2007,brikic_gas_2011}. They are typically of the form $Q_{ij} \appropto  |p_i^2 - p_j^2 + b_{ij}|^{0.5 - 0.55}$, as we used for our gas-flow analysis in \ref{sec:4}. Here the constant $b_{ij}$ describes the effect of gravity in case of height differences between the ends of the pipe, and the proportionality constant describes friction effects \cite{abeysekera_steady_2016, coelho_considerations_2007,brikic_gas_2011}.
    \item Laminar flow (low-velocity flow, thin pipes): A relation between the flow rate through a pipe and the pressure at either end can be calculated directly from the Navier-Stokes equation \cite{batchelor_introduction_2000}, giving the linear relationship $Q_{ij} = a_{ij}(P_i - P_j)$.
    Here $Q_{ij}$ is the flow rate from point $i$ to point $j$, $P_i$ and $P_j$ are the respective pressures at point $i$ and point $j$, and $a_{ij}$ is a constant determined by the dimensions of the pipe and the viscosity of the fluid. BP is directly applicable in this case.
    \item Incompressible fluids (such as liquids or low-pressure gases).
    Fluids that can be treated as incompressible comprise gas flow in municipal distribution networks, water flow in water works, air flow in ventilation systems in buildings or district heating, or cooling systems.
    Incompressible fluid means that the pressure drop is negligible and the density remains approximately constant.
    In common to fluid networks, when the resistance in the pipe depends on the flow, the problem is nonlinear\cite{walski_history_2006, williams_hydraulic_1905, brown_history_2002}. As in \ref{sec:4}, one can use an iterative procedure combined with our BP algorithm to calculate the distribution of fluid flow through the pipes.
\end{itemize}

\section{Benchmark networks}\label{app: benchmark_networks}
Here we discuss the IEEE and Gaslib benchmark networks, which we have used to compare the efficiency of BP on the different factor graphs.
As the benchmark networks represent real networks, their structure and parameters
are highly heterogenous and suitable for testing the applicability of algorithms to practical computations.\\

\begin{figure}
    \centering
    \includegraphics[trim = 300 65 100 290,clip, width = 0.9 \textwidth, angle = -0.4]{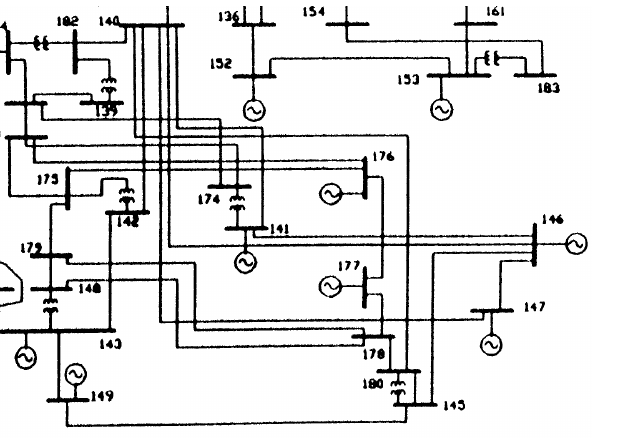}
    \caption{A section of the IEEE-300 network (the network is based on the power grid of the Midwestern USA), from \cite{christie_power_1993}. In our work we only distuingish between the buses (vertices), here represented as bars to which links connect, and the links between the buses.
    The full IEEE-300 network has 300 vertices and 411 links.}
    \label{fig: IEEE-300}
\end{figure}

{\bf The IEEE-300 network.}
The IEEE benchmark systems are a set of networks that
were designed to resemble simplified versions of realistic
power networks \cite{christie_power_1993}. These are commonly used in the power
systems engineering literature to test the applicability of algorithms related to power flow.
Systems of varying sizes are available on \cite{christie_power_1993}.
Here we used the IEEE-300 network, 'developed by the IEEE Test Systems Task Force under the direction of Mike Adibi in 1993' \cite{christie_power_1993}.
The network consist of 300 buses (vertices) and 411 links, and contains 112 loops.
The IEEE-grids are realistic (they are based on the power system in Midwestern USA) and therefore highly heterogenous
both in the topology and in the parameters; a section of the IEEE-300 network is shown in Fig.~\ref{fig: IEEE-300}.\\
In principle different power devices such as transformers are contained in the network;
for simplicity we only consider buses and links. The data contains impedances and resistances of each transmission line, as well as the phase angles at each bus, which together fully specify the network in the DC approximation.\\

\begin{figure}
    \centering
    \subfloat[]{\includegraphics[width =0.64\textwidth]{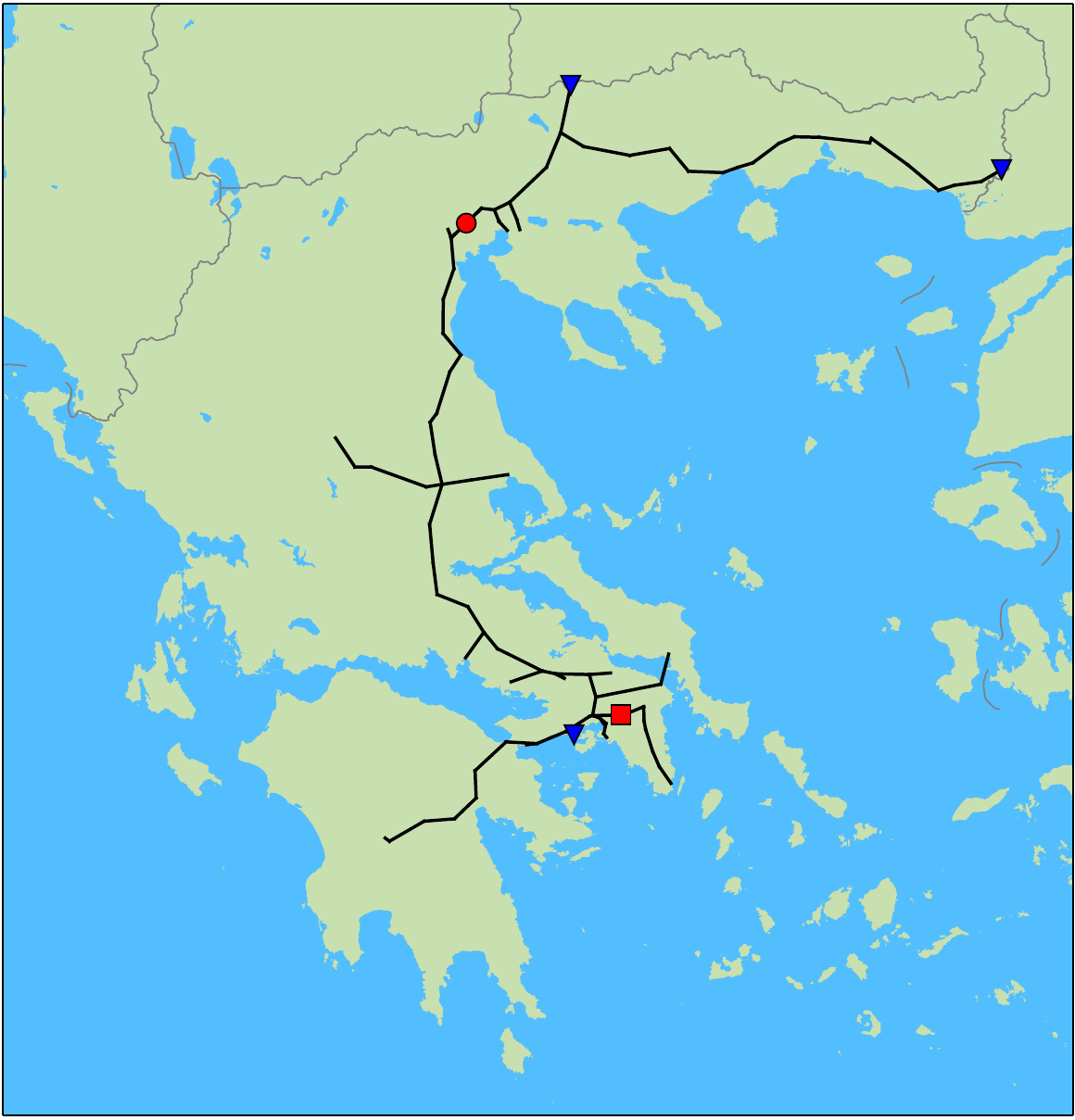}} \\
    \vspace{0.1cm} \hspace{-0.7cm}\subfloat[]{\includegraphics[width = 0.526 \textwidth]{figure_9b-eps-converted-to.pdf}}
    \caption{(a) Topology of the GasLib-134 network, a tree network representing the gas pipeline connection between Greece and Bulgaria. (b) Topology of the Gaslib-40 network. Both figures are from \cite{schmidt_gasliblibrary_2017}.}
    \label{fig: gaslib}
\end{figure}

{\bf The GasLib networks.}
The GasLib networks \cite{schmidt_gasliblibrary_2017} were developed to provide researchers with large and realistic gas benchmark networks, as part of a project funded by the German Federal Ministry of Economics and Technology.
The GasLib-134 network is based on the gas-pipeline network connecting Greece and Bulgaria (shown in Fig~\ref{fig: gaslib}), albeit somewhat distorted to protect sensitive data.
GasLib-134 contains 134 vertices, 133 links and no loops. GasLib-40 contains 40 vertices, 45 links and 6 loops. For simplification we do not consider any devices contained in the network, and only distinguish the vertices and links.
The GasLib data contain the network topology and the gas injections. For GasLib-134 we have used the scenario data from 2011-11-01. Computation of the coefficients in Eq.~\ref{eq15} is less straightforward,
and here we simply set the coefficients of all pipes to the same value (50 for GasLib-134, 500 for GasLib-40), determined such that pressures maintain realistic values (between $40$ and $60$ bar).
Coefficients with a small random component give similar results.
To get results for the steady state analysis in \ref{sec:4} the vertices with known pressure values are the 'source\_1' for GasLib-40 and 'node\_1', 'node\_20' and 'node\_80' for GasLib-134.
Combined with the injection at all other vertices, the systems are then fully determined.

\section{Algorithmic details} \label{sec: pseudocode}
\subsection{State estimation}\label{sec: stateest_pseudocode}
In the state estimation problem of Section~\ref{sec:3} we have available the topology of the power grid, a set of measurements $\{z_{gi}\}$, $\{z_{fij}\}$ and $\{z_{\theta_i}\}$, and their associated variances $\{\sigma^2_{gi}\}$, $\{\sigma^2_{fij} \}$ and $\{\sigma^2_{\theta_i}\}$. Here the index $i$ denotes the vertices of the power grid, and $(ij)$ denotes the links of the power grid. In our case we have  generated the measurements ourselves according to the procedure in Section~\ref{sec:3}.
We furthermore have access to the set of susceptances $\{B_{ij}\}$, such that the flows are related to the phase angles as $f_{ij} = B_{ij}(\theta_i - \theta_j)$.
Our goal is then to accurately determine the flows, based on the Bayes' theorem (Eq.~\ref{eq: dist_factorization}). Let us define for all links $(ij)$ the constants $c_{fij|i} = B_{ij} $, $c_{fij|j} = -B_{ij}$ and $c_{gi|j} = -B_{ij}$, and for all vertices $i$ the constants $c_{gi|i} =  \sum_{j \in N(i)}B_{ij}$. For a given set of measurements Bayes' theorem gives a probability $P(\bm{\theta})$ that the phase angles have values $\bm{\theta}$ (the posterior distribution $P(\bm{\theta}|\bm{z})$ from Eq.~\ref{eq: dist_factorization})
as:
\begin{eqnarray}
 P(\bm{\theta}\vert\bm{z}) \equiv  P(\bm{\theta}) &=& \prod_i H_i(\theta_i, \theta_j: j \in N(i)) \nonumber\\
 &\times& \prod_{(ij)} H_{(ij)}(\theta_i,\theta_j) \, \\
    H_{(ij)}(\theta_i,\theta_j) &\propto& \exp\Big(-(z_{fij} - c_{fij|i}\theta_i - c_{fij|j}\theta_j)^2/2\sigma^2_{fij}\Big) \,, \nonumber \\ \label{eq: form_v_Hij}
\end{eqnarray}
\begin{eqnarray} \label{eq: form_v_Hi}
    H_i(\theta_i, \theta_j: j \in N(i))   \propto  \exp\Big(-(z_{\theta_i} - \theta_i)^2/2\sigma_{\theta_i}^2&\Big)&  \nonumber \\
       \times \exp\Big(-(z_{gi} - c_{gi|i}\theta_i - \sum_{j \in N(i)}c_{gi|j}\theta_j)^2/2\sigma_{gi}^2&\Big)&\,.
\end{eqnarray}
This distribution is of the form of Eq.~\ref{eq: basicP}, and hence we can use Belief Propagation on the different factor graphs $F_v$ and $F_c$.
To use the message passing equations (Eqs.~\ref{eq: BP1}-\ref{eq: BP4}), we need the set of variable nodes $x_I$, the set of factor nodes $a$ and the associated factor functions $W_a(\bm{x}_a)$.
Summarizing the main text, these are given in Tab.~\ref{tab: factor_graphs}.
\begin{table*}
{\renewcommand{\arraystretch}{1.2}
\begin{tabular}{c c c}
    \hline
    Factor graph & Variable nodes $x_I$ & Factor nodes $W_a(\bm{x}_a)$\\
    \hline
    $F_v$ &  \begin{tabular}{c} for each vertex $i$: \\ $\theta_i$  \end{tabular}& \begin{tabular}{c} for each vertex $i$: \\ $H_i(\theta_i, \theta_j: j \in N(i))$ \\ for each link $(ij)$: \\ $H_{(ij)}(\theta_i,\theta_j)$\end{tabular} \\
    \hline
    $F_c$ & \begin{tabular}{c} for each link $(ij)$: \\ $(\theta_{ij}^c, \theta_{ji}^c)$ \end{tabular} & \begin{tabular}{c} for each vertex $i$: \\  $\int \mathrm{d} \theta_i \, \Big\{H_i(\theta_i, \theta^c_{ji}: j \in N(i)) \prod_{j \in N(i)} \delta(\theta^{c}_{ij} - \theta_i) \Big\}$ \\ for each link $(ij)$: \\\hspace{0.2cm}$H_{(ij)}(\theta^c_{ij},\theta_{ji}^c)$ \end{tabular} \\
    \hline
\end{tabular}}
\caption{The factor graphs $F_v$ and $F_c$ specified for the power grid state estimation problem.}
\label{tab: factor_graphs}
\end{table*}
For the flow-only approximation, which ignores Kirchhoff's second law and the phase angle measurements, the posterior distribution described by Bayes' theorem is:
\begin{eqnarray}
    P_f(\bm{f}\vert\bm{z}) \equiv P_f(\bm{f}) &= &\Big[ \prod_{i} H_i(\sum_{j \in N(i)}f_{ij})\Big] \times \Big[ \prod_{(i,j)} H_{ij}(f_{ij}) \Big] \,,\nonumber \\ && \\
    H_i(\sum_{j \in N(i)}f_{ij}) &\propto& \exp\Big(-(z_{gi} - \sum_{j \in N(i)} f_{ij})^2/2\sigma_{gi}^2 \Big)\,, \label{eq: form_f_Hi}\\
    H_{ij}(f_{ij}) &\propto& \exp\Big(-(z_{fij} - f_{ij})^2/2\sigma^2_{fij} \Big) \label{eq: form_f_Hij} \,.
\end{eqnarray}
This distribution is of the form of Eq.~\ref{eq: Pflow}. Following the main text, the factor graph $F_f$ is given in Tab.~\ref{tab: F_f}.  \\ \\
\begin{table*}
    {\renewcommand{\arraystretch}{1.2}
    \begin{tabular}{c c c}
        \hline
        Factor graph & Variable nodes $x_I$ & Factor nodes $W_a(\bm{x}_a)$\\
        \hline
        $F_f$ &  \begin{tabular}{c} for each link $(ij)$: \\ $f_{ij}$  \end{tabular}& \begin{tabular}{c} for each vertex $i$: \\ $H_i(\sum_{j \in N(i)}f_{ij})$ \\ for each link $(ij)$: \\ $H_{ij}(f_{ij})$\end{tabular} \\
        \hline
    \end{tabular}}
    \caption{The factor graph $F_f$ specified for the power grid state estimation problem.}
    \label{tab: F_f}
\end{table*}
Using these factor graphs, we can use the BP Eqs.~\ref{eq: BP1}-\ref{eq: BP4} to calculate the marginals we are interested in. We focus on estimating the flows $f_{ij} = B_{ij}(\theta_i - \theta_j)$, so for $F_v$ this is $\{b_{Hij}(\theta_i, \theta_j) \}$ where the index $H_{ij}$ denotes the factor node at the link $(ij)$. For $F_c$ this is calculated from the copies as $b_{(ij)}(\theta_i, \theta_{j}) = b_{(ij)}(\theta^c_{ij}, \theta^c_{ji})$, where the index $(ij)$ denotes the variable node at the link $(ij)$. For $F_f$ the relevant marginals are $\{b_{(ij)}(f_{ij})\}$, where $(ij)$ is again the variable node at the link $(ij)$.
Due to the fact that all $H_i$ and $H_{ij}$ are Gaussian, all of the messages remain Gaussian,
such that we can summarize them by their mean and covariance as $m_{I \rightarrow a}(x_I) = N(\mu_{I \rightarrow a}, K_{I \rightarrow a})$ and $m_{a \rightarrow I}(x_I) = N(\mu_{a \rightarrow I}, K_{a \rightarrow I})$ for all variable nodes $I$ and factor nodes $a$ connected on the factor graph (that is, for which $x_I \in \bm{x}_a$).
Together with the topology of the factor graph and the expressions for the functions $W_a(\bm{x}_a)$ from tables \ref{tab: factor_graphs} and \ref{tab: F_f},
we can plug this into the BP Eqs.~\ref{eq: BP1}-\ref{eq: BP3} to obtain the algorithm in terms of
products and integrals over Gaussians, therefore requiring only linear algebra for the means and (co-)variances of the messages. It turns out that for $H_i$ and $H_{ij}$ of the form of Eqs.~\ref{eq: form_v_Hij}-\ref{eq: form_v_Hi} and Eqs.~\ref{eq: form_f_Hi}-\ref{eq: form_f_Hij} the updates can be calculated particularly efficiently (that is, not requiring explicit inversion of any matrices larger than 2x2). The pseudocode for the resulting algorithms is given in the supplementary material.

%
%
\subsection{Steady state analysis}\label{sec: steadystate_pseudocode}
For the steady state analysis problem of \ref{sec:4} we have access to the topology of the gas network, to the known values of the squared pressures $\overline{\bm{v}} = \{\overline{v}_{i} \}$ at some vertices $i$, the known values of injection $\overline{\bm{g}} = \{ \overline{g}_i \}$ at some vertices $i$, and to the coefficients $\{a_{ij}\}$ determining the flow $Q_{ij}(v_i,v_j)=a_{ij} \mbox{sgn}(v_i-v_j)(v_i-v_j)^{0.5}$. Here the variables $v_i$ correspond to the square of the pressure $p_i^2$ at the vertex $i$.
From \ref{sec:4}, at each iteration of the modified GN-algorithm we have a distribution $P(\bm{v}) \propto \exp(-C(\bm{v}))$, with $C(\bm{v})$ as in Eq.~\ref{eq: cost_function} (the resulting distribution is Gaussian, implying its mean minimizes $C(\bm{v})$). The resulting distribution $P(\bm{v})$ is of the same form as those for the state estimation problem (Eqs.~\ref{eq: form_v_Hij}-\ref{eq: form_v_Hi}), only with different coefficients. Exactly the same BP algorithm can thus be used, only replacing $\{\theta_i\}$ with $\{v_i\}$ = $\{p_i^2\}$, and replacing the values of the coefficients as:
\begin{eqnarray}
z_{gi} =\begin{cases}
        \overline{g}_i - \sum_{j \in N(i)}\Big(Q_{ij}(v_i^\ast, v_j^\ast) - \frac{\partial Q_{ij}}{\partial v_i}\biggr\rvert_{v_i^\ast, v_j^\ast} v_i^\ast & \! \! \! \! \! \! \! \!- \frac{\partial Q_{ij}}{\partial v_j}\biggr\rvert_{v_i^\ast, v_j^\ast} v_j^\ast\Big)  \\
        &\text{if} \quad \overline{g}_i \in \bm{\overline{g}}\,, \\
         0&  \text{else} \,,
    \end{cases} \nonumber
\end{eqnarray}
\begin{eqnarray}
    \sigma^2_{gi} &=& \begin{cases}
        1   &\text{if} \quad \overline{g}_i \in \bm{\overline{g}}\,, \\
        10^8 & \text{else} \,,\end{cases}  \nonumber\\
     z_{vi} &=& \begin{cases}
            \overline{v}_i   &\text{if} \quad \overline{v}_i \in \bm{\overline{v}}\,, \\
             0 & \text{else} \,,
    \end{cases} \nonumber \\
      \sigma^2_{vi} &=& \begin{cases}
            1   &\text{if} \quad \overline{v}_i \in \bm{\overline{v}}\,, \\
            10^8 & \text{else} \,,
    \end{cases} \nonumber \\
    c_{gi|i} &=& \begin{cases}
        \sum_{j \in N(i)}\frac{\partial Q_{ij}}{\partial v_i}\biggr\rvert_{v_i^\ast, v_j^\ast}   &\text{if} \quad \overline{g}_i \in \bm{\overline{g}}\,, \\
         1 & \text{else} \,, \end{cases} \nonumber \\
    z_{fij} &=& 0\,,  \nonumber\\
    c_{fij|i} &=& -c_{fij|j} = 1\,, \nonumber \\
    \sigma^2_{fij} &=& 10^8 \,.  \nonumber\\
     c_{gi|j} &=& \begin{cases}
        \frac{\partial Q_{ij}}{\partial v_j}\biggr\rvert_{v_i^\ast, v_j^\ast}   &\text{if} \quad \overline{g}_i \in \bm{\overline{g}}\,, \\
    1 & \text{else} \,. \end{cases}  \nonumber
\end{eqnarray}
Here the value $10^8$ is simply a very high number implementing the absence of any knowledge of the value of the corresponding variable (a value much lower than $10^8$ can lead to inaccuracies, while a much higher value can lead to numerical over- or underflow). The linearization point $\bm{v}^\ast$ can be directly obtained from a guess $\bm{Q}^\ast$ (itself obtained from the previous BP iteration), the precise algorithm is given in the supplementary material.
For the gas network we can also use a flow-only approximation, giving a distribution of the form of Eqs.~\ref{eq: form_f_Hi}-\ref{eq: form_f_Hij}, with the flow $f_{ij}$ replaced by $Q_{ij}$, and the coefficients: 
\begin{eqnarray}
    \quad z_{gi} &=& \begin{cases}
        \overline{g}_i &\text{if} \quad \overline{g}_i \in \bm{\overline{g}}\,, \\
         0 & \text{else} \,,
    \end{cases} \nonumber \\
   \sigma^2_{gi} &=& \begin{cases}
        1   &\text{if} \quad \overline{g}_i \in \bm{\overline{g}}\,, \\
        10^8 & \text{else} \,,\end{cases} \nonumber\\
      z_{fij} &=0& \,, \nonumber \\
        \sigma^2_{fij} &=& 10^8 \,. \nonumber
\end{eqnarray}

\bibliographystyle{iopart-num}
\bibliography{main}

\end{document}


\title{Supplementary Material to 'Belief propagation for supply networks: Efficient clustering of their factor graphs'}

\author{Tim Ritmeester\thanksref{e1,addr1}
        \and
        Hildegard Meyer-Ortmanns\thanksref{e2,addr1} 
}

\thankstext{e1}{e-mail: t.ritmeester@jacobs-university.de}
\thankstext{e2}{e-mail: h.ortmanns@jacobs-university.de}

\institute{Physics and Earth Sciences,
  Jacobs University Bremen,  P.O. Box 750561, 28725 Bremen, Germany \label{addr1}
}
\date{}
\maketitle

\vspace{-1cm}
\section{Pseudocode}
Our clustering method is formulated in the main text in terms of factor graphs, as the factor graph can be used for a variety of adaptations to the basic BP-algorithm, as discussed in the main text. In our test cases we used the basic BP-algorithm (section `sketch of the BP algorithm'), which we found to perform well. Inserting the factor graphs $F_v$ and $F_f$ into the BP update equations gives Gaussian integrals and products, which can be calculated by matrix inversions and products of the (co-)variance and means of the messages. The same is true for $F_c$, for which the BP update equations simplify after integrating over the delta functions. Doing so gives the following `black-box' update equations for messages between factors nodes corresponding to $H_i$ (denoted by the index `$i$') and variable nodes corresponding to $v_i, v_j$ (denoted by the index `$(ij)$'):
\begin{alignat}{1}
    m_{(ij) \rightarrow i}^{t+1}(v_i, v_j) &= \frac{b_{(ij)}^{t}(v_i, v_j) }{ m_{i \rightarrow (ij)}^t(v_j, v_i)} \,,  \\
    m_{i \rightarrow (ij)}^{t+1}\big(v_i, v_j \big) &= \int  \Big[ \hspace{-0.5cm}\prod_{v_k: k \in N(i) \setminus  j} \hspace{-0.0cm} \mathrm{d}v_k \Big]\frac{b^{t}_{i}(v_i, \{v_j: j \in N(i)\})}{m^t_{(ij) \rightarrow i}(v_j, v_i) } \\
    b_{(ij)}^{t+1}(v_i, v_j) &= H_{(ij)}\big(v_j, v_i \big) \nonumber \\
    &\times m_{j \rightarrow (ij)}^t(v_j, v_i) \times m_{i \rightarrow (ij)}^t(v_j, v_i)\,, \nonumber \\
    b^{t+1}_{i}(v_i, \{v_j: j \in N(i)\}) &=  H_i\big(v_i, v_j:j\in N(i)\big) \Big] \times \Big[ \hspace{-0.5cm} \prod_{(v_k): k \in N(i) \setminus  j}  \hspace{-0.0cm} m_{(ik) \rightarrow i}(v_i,v_k)  \Big] 
\end{alignat}
For all three factor graphs the message updates can be computed by matrix inversion and multiplication. For the specific distributions we considered (appendix `Algorithmic details'), the algorithm can be made somewhat more efficient by symbolically performing the inversions and multiplications, leading to the pseudocode given below.
\captionof{algorithm}{Algorithm for $F_f$}
\begin{algorithmic}[1]
    \LeftComment{Set the initial variance (${K}$) and mean (${\mu}$) of the messages.}
    \LeftComment{Messages from/to variable nodes are denoted with indices $(ij)$.}
    \LeftComment{Messages from/to factor nodes are denoted with indices $H_i$ (the factor nodes $H_{ij}$ are leaf nodes, the corresponding messages do not have to be calculated explicitly).}
    \ForAll{$i$}
    \ForAll{$j \in N(i)$}
    \State ${K}_{Hi \rightarrow (ij)} \gets 10^6$
    \State ${\mu}_{Hi \rightarrow (ij)} \gets  0 $
    \State ${K}_{(ji) \rightarrow Hi} \gets 10^6 $
    \State ${\mu}_{(ji) \rightarrow Hi} \gets 0$
    \EndFor
    \EndFor
    \LeftComment{Updating procedure:}
    \Repeat:
    \ForAll{$i$}
        \ForAll{$j \in N(i)$}
        \State  $K_{(ji) \rightarrow Hi} \gets \Big(\big({K}_{Hj \rightarrow (ji)}\big)^{-1} + 1/\sigma_{fji}^2 \Big)^{-1}$
        \State $\mu_{(ji) \rightarrow Hi} \gets{K}_{(ji) \rightarrow Hi}\Big(\big({K}_{Hj \rightarrow (ji)}\big)^{-1}{\mu}_{Hj \rightarrow (ji)} + z_{fji}/\sigma_{fji}^2 \Big)$
        \EndFor
        \EndFor
    \ForAll{$i$}
        \State $\gamma \gets \sum_{Hj \in N(i)} K_{(ji) \rightarrow Hi}$
        \State $\epsilon \gets \sum_{Hj \in N(i)} \mu_{(ji) \rightarrow Hi}$

        \ForAll{$j \in N(i)$}
        \State$\gamma_{(ij)} \gets \gamma - K_{(ji) \rightarrow Hi}$
        \State $\epsilon_{(ij)} \gets \epsilon - \mu_{(ji) \rightarrow Hi}$
        \State $K_{Hi \rightarrow (ij)} \gets (\sigma_{gi}^2 + \gamma_{(ij)})$
        \State $\mu_{Hi \rightarrow (ij)} \gets  (z_{gi} + \epsilon_{(ij)})$
        \EndFor
        \EndFor
    \Until{Converged.}
    \LeftComment{Collect results:}

    \ForAll{$i$}
    \ForAll{$j \in N(i)$}
    \State ${K}_{j \rightarrow (ij)} \gets {K}_{j \rightarrow (ji)}$
    \State ${\mu}_{j \rightarrow (ij)} \gets -{\mu}_{j \rightarrow (ji)}$
    \State ${K}_{(ij)} \gets \Big(\big({K}_{i \rightarrow (ij)}\big)^{-1} + \big({K}_{j \rightarrow (ij)}\big)^{-1}\Big)^{-1}$
    \State ${\mu}_{(ij)} \gets {K}_{(ij)} \Big(\big({K}_{i \rightarrow (ij)}\big)^{-1} {\mu}_{i \rightarrow (ij)} + \big({K}_{j \rightarrow (ij)}\big)^{-1} {\mu}_{j \rightarrow (ij)}\Big)$
    \EndFor
    \EndFor
    \LeftComment{Output: $b_{(ij)}(f_{ij}) = N({\mu}_{(ij)}, {K}_{(ij)})$ for each link $(ij)$.}
    \label{alg: F_f}
\end{algorithmic}
\captionof{algorithm}{Algorithm for $F_c$}
\begin{algorithmic}[1]
    \LeftComment{Set the initial covariance ($\bm{K}$) and mean ($\bm{\mu}$) of the messages.}
    \LeftComment{Messages from/to variable nodes are denoted with indices $(ij)$}
    \LeftComment{Messages from/to factor nodes are denoted with indices $H_i$ (the factor nodes $H_{ij}$ are leaf nodes, the corresponding messages do not have to be calculated explicitly).}
    \ForAll{$i$}
    \ForAll{$j \in N(i)$}
    \State $\bm{K}_{Hi \rightarrow (ij)} \gets \begin{bmatrix}
                                        10^6 & 0 \\
                                        0 & 10^6
                                        \end{bmatrix}$
    \State $\bm{\mu}_{Hi \rightarrow (ij)} \gets \begin{bmatrix}
                                        0 \\
                                        0
                                        \end{bmatrix}$
    \State $\bm{K}_{(ji) \rightarrow Hi} \gets \begin{bmatrix}
                                        10^6 & 0 \\
                                        0 & 10^6
                                        \end{bmatrix}$
    \State $\bm{\mu}_{(ji) \rightarrow Hi} \gets \begin{bmatrix}
                                        0 \\
                                        0
                                        \end{bmatrix}$
    \EndFor
    \EndFor
    \LeftComment{Updating procedure:}
    \Repeat:
    \ForAll{$i$}
        \ForAll{$j \in N(i)$}
        \State  $\bm{K}_{(ji) \rightarrow Hi} \gets \Big(\big(\bm{K}_{Hj \rightarrow (ji)}\big)^{-1} + \frac{1}{\sigma^2_{fji}}\begin{bmatrix}c_{fji|j}^2 & c_{fji|i} \cdot c_{fji|j}\\ c_{fji|i} \cdot c_{fji|j}& c_{fji|i}^2\end{bmatrix}\Big)^{-1}$
        \State $\bm{\mu}_{(ji) \rightarrow Hi} \gets \bm{K}_{(ji) \rightarrow Hi}\Big(\big(\bm{K}_{Hj \rightarrow (ji)}\big)^{-1}\bm{\mu}_{Hj \rightarrow (ji)} + \frac{z_{fji}}{\sigma^2_{fji}}\begin{bmatrix}  c_{fji|j}\\  c_{fji|i}\end{bmatrix} \Big)$
        \EndFor
        \EndFor
    \ForAll{$i$}
        \ForAll{$j \in N(i)$}
            \State $\begin{bmatrix}
                a_{jj} & a_{ji}\\
                a_{ji} & A_j
            \end{bmatrix} \gets (\bm{K}_{(ji) \rightarrow Hi})^{-1} $
            \State $ \begin{bmatrix} \mu_j \\ \mu_{j i} \end{bmatrix} \gets \bm{\mu}_{(ji) \rightarrow Hi}$
        \EndFor
        \State $\beta \gets -\frac{1}{\sigma_i^2} + \sum_{j \in N(i)} \big(a_{ji}^2/a_{jj} - A_j)$
        \State $\gamma \gets \sum_{j \in N(i)} c_{gi|j}^2/a_{jj}$
        \State $\delta \gets \sum_{j \in N(i)} a_{ji} c_{gi|j}/a_{jj}$
        \State $\epsilon \gets \sum_{j \in N(i)} c_{gi|j} \mu_j/a_{jj}^2$
        \State$\zeta \gets \frac{z_{vi}}{\sigma_i^2} + \sum_{j \in N(i)} \big(\mu_{ji}/A_j - a_{ji}\mu_j/a_{jj}^2 \big)$

        \ForAll{$j \in N(i)$}
        \State $\beta_j \gets \beta - \big(a_{ji}^2/a_{jj} - A_j)$
        \State $\gamma_j \gets \gamma - \frac{c_{gi|j}^2}{a_{jj}}$
        \State $\delta_j \gets \delta - \frac{a_{ji} c_{gi|j}}{a_{jj}}$
        \State $\epsilon_j \gets \epsilon - \frac{c_{gi|j} \mu_j}{ a_{jj}^2}$
        \State $\zeta_j \gets \zeta - \big(\frac{\mu_{ji}}{ A_j} - \frac{a_{ji}\mu_j}{a_{jj}^2} \big)$
        \State $\bm{K}_{Hi \rightarrow (ij)} \gets \begin{bmatrix}
                                                             -1/\beta_j & \frac{c_{gi|i} - \delta_j}{c_{gi|j} \beta_j} \\
                                                            \frac{c_{gi|i} - \delta_j}{c_{gi|j} \beta_j} & \frac{\gamma_j + \sigma_{gi}^2}{c_{gi|j}^2} - \frac{(c_{gi|i} - \delta_j)^2}{(c_{gi|j}^2\beta_j)}
                                                            \end{bmatrix}$
        \State $\bm{\mu}_{Hi \rightarrow (ij)} \gets \begin{bmatrix}
            - \zeta_j/\beta_j  \\
            (z_{gi} - \epsilon_j)/c_{gi|j} + (c_{gi|i} - \delta_j)(\zeta_j)/c_{gi|j}\beta_j
                                                  \end{bmatrix}$
        \EndFor
        \EndFor
    \Until{Converged.}
    \LeftComment{Collect results:}

    \ForAll{$i$}
    \ForAll{$j \in N(i)$}
    \LeftComment{$\text{flip}\Big(\begin{bmatrix} a & b \\ c & d \end{bmatrix}\Big) \equiv \Big(\begin{bmatrix} d &c \\ b & a \end{bmatrix} \Big)$.}
    \LeftComment{$\text{flip}\Big(\begin{bmatrix} a \\ b \end{bmatrix}\Big) \equiv \Big(\begin{bmatrix} b \\ a \end{bmatrix} \Big)$.}
    \State $\bm{K}_{Hj \rightarrow (ij)} \gets \text{flip}\big(\bm{K}_{Hj \rightarrow (ji)}\big)$
    \State $\bm{\mu}_{Hj \rightarrow (ij)} \gets \text{flip}\big(\bm{\mu}_{Hj \rightarrow (ji)}\big)$
    \State $\bm{K}_{(ij)} \gets \Big(\big(\bm{K}_{Hi \rightarrow (ij)}\big)^{-1} + \big(\bm{K}_{Hj \rightarrow (ij)}\big)^{-1}\Big)^{-1}$
    \State $\bm{\mu}_{(ij)} \gets \bm{K}_{(ij)} \Big(\big(\bm{K}_{Hi \rightarrow (ij)}\big)^{-1} \bm{\mu}_{Hi \rightarrow (ij)} + \big(\bm{K}_{Hj \rightarrow (ij)}\big)^{-1} \bm{\mu}_{Hj \rightarrow (ij)}\Big)$
    \EndFor
    \EndFor
    \LeftComment{Output: $b_{(ij)}(\theta_i, \theta_j) = N(\bm{\mu}_{(ij)}, \bm{K}_{(ij)})$ for each link $(ij)$.}
    \label{alg: F_c}
\end{algorithmic}
\captionof{algorithm}{Algorithm for $F_v$}
\begin{algorithmic}[1]
    \LeftComment{Set the initial variance ($K$) and mean ($\mu$) of the messages.}
    \LeftComment{Messages from/to variable nodes are denoted with indices $i$}
    \LeftComment{Messages from/to factor nodes are denoted with indices $H_i$ and $H_{ij}$.}
    \ForAll{$i$}
    \State $K_{i \rightarrow Hi} \gets 10^6$
    \State $\mu_{i \rightarrow Hi} \gets 0$
    \State $K_{Hi \rightarrow i} \gets 10^6$
    \State $\mu_{Hi \rightarrow i} \gets 0$
    \ForAll{$j$}
    \State $K_{i \rightarrow Hij} \gets 10^6$
    \State $\mu_{i \rightarrow Hij} \gets 0$
    \State $K_{Hji \rightarrow i} \gets 10^6$
    \State $\mu_{Hji \rightarrow i} \gets 0$
    \EndFor
    \EndFor
    \LeftComment{Updating procedure:}
    \Repeat:
    \ForAll{$i$}
    \State $K_i \gets \Big(\frac{1}{\sigma_{vi}^2} + (K_{Hi \rightarrow i})^{-1} + \sum_{j \in N(i)}(K_{Hj \rightarrow i})^{-1} + \sum_{j \in N(i)}(K_{Hji \rightarrow i})^{-1}\Big)^{-1}$
    \State \begin{eqnarray*} \hspace{1cm} \mu_i \gets K_i \Big(\frac{z_{vi}}{\sigma_{vi}^2} + (K_{Hi \rightarrow i})^{-1}\mu_{Hi \rightarrow i} + \sum_{j \in N(i)}(K_{Hj \rightarrow i})^{-1}\mu_{Hj \rightarrow i} + \sum_{j \in N(i)}(K_{Hji \rightarrow i})^{-1}\mu_{Hji \rightarrow i}\Big) \end{eqnarray*}
     \State $K_{i \rightarrow Hi} \gets \Big(\big(K_i\big)^{-1} - \big(K_{Hi \rightarrow i}\big)^{-1}\Big)^{-1}$
    \State $\mu_{i \rightarrow Hi} \gets K_{i \rightarrow Hi}\Big(\big(K_i\big)^{-1}\mu_i - \big(K_{Hi \rightarrow i}\big)^{-1}\mu_{Hi \rightarrow i} \Big)$

    \ForAll{$j \in N(i)$}
     \State $K_{i \rightarrow Hj} \gets \Big(\big(K_i\big)^{-1} - \big(K_{Hj \rightarrow i}\big)^{-1}\Big)^{-1}$
    \State $\mu_{i \rightarrow Hj} \gets K_{i \rightarrow Hj}\Big(\big(K_i\big)^{-1}\mu_i - \big(K_{Hj \rightarrow i}\big)^{-1}\mu_{Hj \rightarrow i} \Big)$
    \State $K_{i \rightarrow Hij} \gets \Big(\big(K_i\big)^{-1} - \big(K_{Hji \rightarrow i}\big)^{-1}\Big)^{-1}$
    \State $\mu_{i \rightarrow Hij} \gets K_{i \rightarrow Hij}\Big(\big(K_i\big)^{-1}\mu_i - \big(K_{Hij \rightarrow i}\big)^{-1}\mu_{Hij \rightarrow i} \Big)$
    \EndFor
    \EndFor
        \ForAll{$i$}
        \State $\gamma \gets c_{gi|i}^2 K_{i \rightarrow Hi} + \sum_{j \in N(i)} c_{gi|j}^2 K_{j \rightarrow Hi}$
        \State $\epsilon \gets c_{gi|i} \mu_{i \rightarrow Hi} + \sum_{j \in N(i)}c_{gi|j} \mu_{j \rightarrow Hi}$
        \State $K_{Hi\rightarrow i} \gets (\sigma_{gi}^2 + \gamma -  c_{gi|i}^2 K_{i \rightarrow Hi})/c_{gi|i}^2 $
        \State $\mu_{Hi \rightarrow i} \gets (z_{gi} - \epsilon +  c_{gi|i} \mu_{i \rightarrow Hi})/c_{gi|i}$

        \ForAll{$j \in N(i)$}
        \State$\gamma_j \gets \gamma - c_{gi|j}^2 K_{j \rightarrow Hi}$
        \State $\epsilon_j \gets \epsilon - c_{gi|j} \mu_{j \rightarrow Hi}$
        \State $K_{Hi\rightarrow j} \gets (\sigma_{gi}^2 + \gamma_j)/c_{gi|j}^2 $
        \State $\mu_{Hi \rightarrow j} \gets (z_{gi} - \epsilon_j)/c_{gi|j}$
        \State $K_{Hji\rightarrow i} \gets (\sigma_{fji}^2 + c_{fji|j}^2 K_{j \rightarrow Hji})/c_{fji|i}^2 $
        \State $\mu_{Hji \rightarrow i} \gets (z_{fji} - c_{fji|j} \mu_{j \rightarrow Hji})/c_{fji|i}$
        \EndFor
        \EndFor

    \Until{Converged.}
    \LeftComment{Collect results:}
    \ForAll{$(ij)$}
    \State $K_{Hij} \gets \Big(\text{diag}\big((K_{i \rightarrow Hij})^{-1}, (K_{j \rightarrow Hji})^{-1}\big) + \frac{1}{\sigma^2_{fij}}\begin{bmatrix}c_{fij|i}^2 & c_{fij|j} \cdot c_{fij|i}\\ c_{fij|j} \cdot c_{fij|i}& c_{fij|j}^2\end{bmatrix}\Big)^{-1}$
    \State $\mu_{Hij} \gets K_{Hij}\Big( \begin{bmatrix} (K_{i \rightarrow Hij})^{-1}\mu_{i \rightarrow Hij} \\ (K_{j \rightarrow Hji})^{-1}\mu_{j \rightarrow Hji} \end{bmatrix}  + \frac{z_{fij}}{\sigma^2_{fij}}\begin{bmatrix}c_{fij|i} \\ c_{fij|j} \end{bmatrix}\Big)$
    \EndFor
    \LeftComment{Output: $b_{Hij}(\theta_i, \theta_j) = N(\bm{\mu}_{(ij)},\bm{K}_{(ij)})$ for each link $(ij)$.}
    \label{alg: F_v}
\end{algorithmic}
\captionof{algorithm}{Algorithm to turn a guess $\bm{Q}^\ast$ into a guess $\bm{v}^\ast$.}\label{euclid}
\begin{algorithmic}[1]
\LeftComment{Initialize a dictionary for corresponding vertex variables $\bm{v}^\ast$.}
\State \textit{dict}   done $\gets$ \{($i$: False)    $\forall i$\}
\State \textit{dict}   $\bm{v}^\ast \gets$ \{($i$: $0$)    $\forall i$\}
\ForAll{$\overline{v}_i \in \bm{\overline{v}}$}
    \State $\bm{v}^\ast$[$i$] $\gets$ $\overline{v}_i$
    \State done[$i$] $\gets$ True
\EndFor
\Repeat
\ForAll{$i$ such that done[$i$] == True}
\ForAll{$j \in N(i)$ such that done[$j$] == False}
\State $v_j \gets \text{solve}\big(Q_{ij}(\bm{v}^\ast$[$i$],$v_j) = Q_{ij}^\ast \big)$
\State $\bm{v}^\ast$[$j$] $\gets v_j$
\State done[$j$] $\gets$ True
\EndFor
\EndFor
\Until{done[$i$] == True    $\forall i$}
\LeftComment{Output: the guess $\bm{v}^\ast$.}
\label{alg: Q_to_v}
\end{algorithmic}